\documentclass[10pt,journal,compsoc]{IEEEtran}

\usepackage{amsmath,amssymb,amsfonts}
\usepackage{graphicx}
\usepackage{textcomp}
\def\BibTeX{{\rm B\kern-.05em{\sc i\kern-.025em b}\kern-.08em
    T\kern-.1667em\lower.7ex\hbox{E}\kern-.125emX}}
    
\usepackage{fancyhdr}
\usepackage{soul,framed}
\usepackage{xspace}
\usepackage{array}
\usepackage{eqparbox}
\usepackage{url}
\usepackage{longtable}
\usepackage{lipsum}
\usepackage{blindtext}
\usepackage{makecell}
\usepackage{mathtools}
\usepackage{commath}
\usepackage{multirow}
\usepackage{tabularx}
\usepackage[normalem]{ulem}
\usepackage{xspace}
\usepackage{algorithm}
\usepackage{algpseudocode}
\usepackage{amsfonts}
\usepackage{placeins}
\usepackage{amssymb}
\usepackage{multirow}
\usepackage{tabularx}
\usepackage{caption}
\usepackage{subcaption}
\usepackage{pifont}
\usepackage[table, dvipsnames]{xcolor}
\usepackage{xspace}
\usepackage{algorithm}
\usepackage{algpseudocode}
\usepackage{amsfonts}
\usepackage{placeins}
\usepackage{booktabs}
\def\BibTeX{{\rm B\kern-.05em{\sc i\kern-.025em b}\kern-.08em
    T\kern-.1667em\lower.7ex\hbox{E}\kern-.125emX}}

\usepackage{hyperref}

\definecolor{darkgreen}{RGB}{119,185,0}

\ifCLASSOPTIONcompsoc
  \usepackage[nocompress]{cite}
\else
  \usepackage{cite}
\fi

\begin{document}

\title{Exploring the Boundaries of Semi-Supervised Facial Expression Recognition using In-Distribution, Out-of-Distribution, and Unconstrained Data}

\author{Shuvendu~Roy,~\IEEEmembership{Student Member,~IEEE,}
        and~Ali~Etemad,~\IEEEmembership{Senior Member,~IEEE}
\IEEEcompsocitemizethanks{\IEEEcompsocthanksitem S. Roy and A. Etemad are with the Department of ECE and Ingenuity Labs Research Institute, Queen's University, Kingston, Canada.
\protect\\
E-mail: shuvendu.roy@queensu.ca; ali.etemad@queensu.ca
} }

\IEEEtitleabstractindextext{
\begin{abstract}
Deep learning-based methods have been the key driving force behind much of the recent success of facial expression recognition (FER) systems. However, the need for large amounts of labelled data remains a challenge. Semi-supervised learning offers a way to overcome this limitation, allowing models to learn from a small amount of labelled data along with a large unlabelled dataset. While semi-supervised learning has shown promise in FER, most current methods from general computer vision literature have not been explored in the context of FER. In this work, we present a comprehensive study on 11 of the most recent semi-supervised methods, in the context of FER, namely Pi-model, Pseudo-label, Mean Teacher, VAT, UDA, MixMatch, ReMixMatch, FlexMatch, CoMatch, and CCSSL. Our investigation covers semi-supervised learning from in-distribution, out-of-distribution, unconstrained, and very small unlabelled data. Our evaluation includes five FER datasets plus one large face dataset for unconstrained learning. Our results demonstrate that FixMatch consistently achieves better performance on in-distribution unlabelled data, while ReMixMatch stands out among all methods for out-of-distribution, unconstrained, and scarce unlabelled data scenarios. 
\textcolor{black}{
Another significant observation is that with an equal number of labelled samples, semi-supervised learning delivers a considerable improvement over supervised learning, regardless of whether the unlabelled data is in-distribution, out-of-distribution, or unconstrained.}
We also conduct sensitivity analyses on critical hyper-parameters for the two best methods of each setting. To facilitate reproducibility and further development, we make our code publicly available at: \href{https://github.com/ShuvenduRoy/SSL\_FER\_OOD}{github.com/ShuvenduRoy/SSL\_FER\_OOD}.

\end{abstract}

}

\maketitle
\IEEEdisplaynontitleabstractindextext

\IEEEpeerreviewmaketitle

\section{Introduction}

Facial Expression Recognition (FER) \cite{kolahdouzi2022facetoponet,kolahdouzi2021face} is a critical application of computer vision that enables computers to identify and understand human expressions, with applications ranging from health care \cite{tokuno2011usage,thrasher2011mood, sanchez2013inferring} to intelligent vehicles \cite{leng2007experimental}. Deep learning methods have been the driving force behind most of the recent successes in FER. However, one of the major barriers to further improvement of deep learning-based FER is the need for large-scale labelled data. To this end, semi-supervised learning (SSL) has shown immense promise as a solution for improving performance while leveraging minimal supervision.
\begin{figure}[t]
\centering
\includegraphics[width=0.49\textwidth]{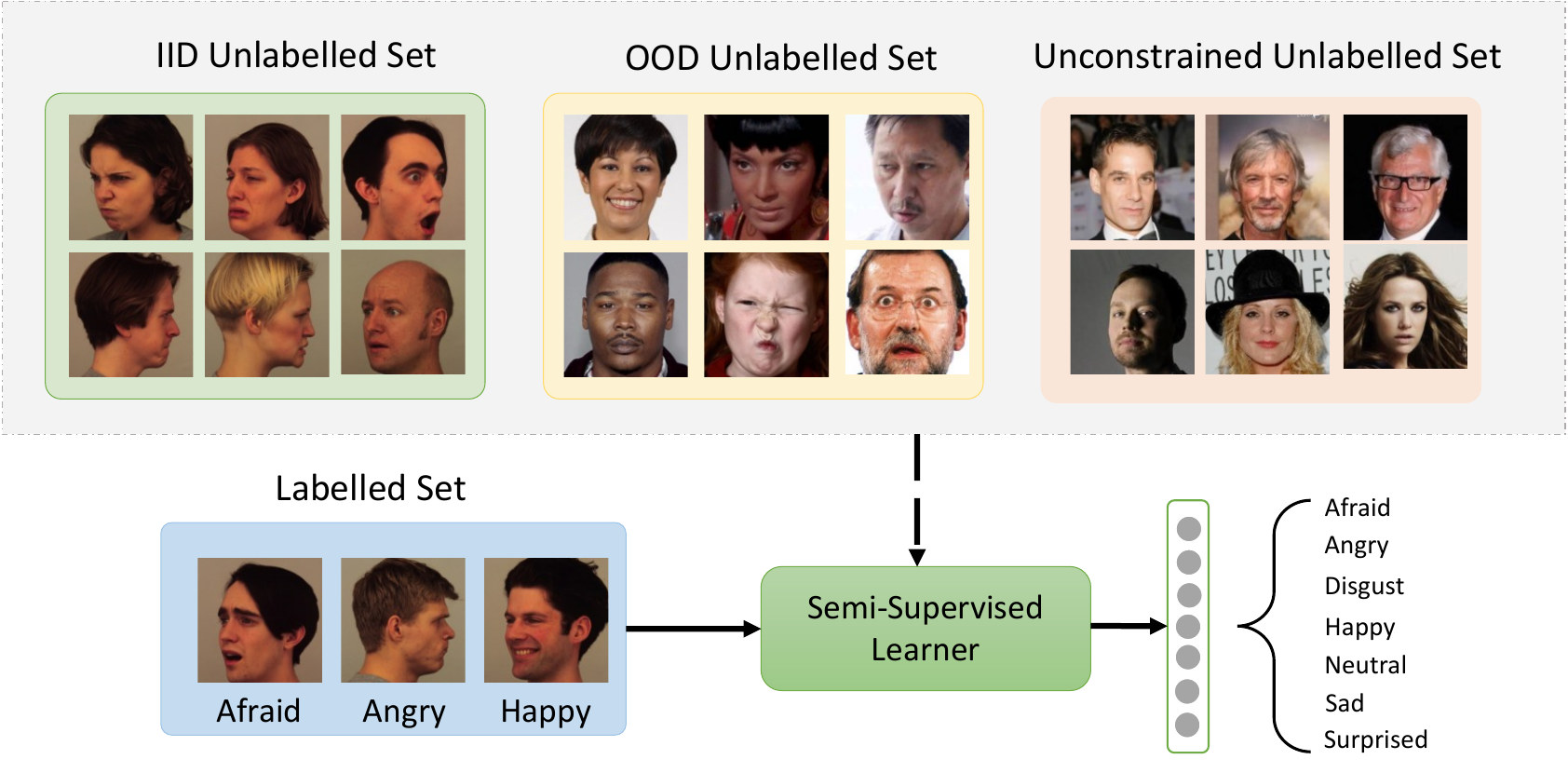}
\caption{Semi-supervised FER under ID, OOD, and unconstrained unlabelled data scenarios.}
\label{fig:bannar}
\end{figure}

\textcolor{black}{
To tackle the problem of label scarcity, semi-supervised methods learn from a small amount of labelled data in conjunction with large amounts of unlabelled data.
Depending on the relationship between the unlabelled and labelled data, there are three well-established forms of SSL: in-distribution (ID) \cite{pseudo_labels,vat,uda,fixmatch,remixmatch}, out-of-distribution (OOD) \cite{yoshihashi2019classification, guo2020safe}, and unconstrained \cite{UnMixMatch,auxmix} SSL. The ID SSL category assumes that the unlabelled data comes from the same distribution as the labelled data. However, in a practical application, this assumption is hard to satisfy or verify when collecting a sizeable unlabelled dataset. Consequently, more recent works have shifted focus toward more realistic data scenarios, including OOD and unconstrained SSL, that offer greater flexibility and potential for real-world applications. In OOD SSL, the unlabelled set contains samples from the same classes as the labelled set but comes from a different source and, therefore, has a different data distribution. On the other hand, unconstrained SSL assumes (1) that the unlabelled data is OOD, and (2) that the unlabeled set can contain samples that belong to classes that are not necessarily the same as those in the labelled set. This makes the unconstrained setting the most practical scenario for collecting large amounts of unlabelled data and scaling up semi-supervised learning \cite{UnMixMatch}. 
}

In the context of FER, we identify the following open research problems regarding SSL: (1) While a few prior studies have explored the ID SSL in the context of FER, no prior works have explored the OOD SSL or unconstrained SSL in FER. (2) Many recent prominent methods originally proposed in general computer vision literature have not been explored in this context. In our recent work \cite{acii2022}, we investigated and benchmarked the performance of several popular semi-supervised learning methods for FER. However, our previous study only focused on the ID SSL.

In this paper, we extend our previous study \cite{acii2022} by exploring semi-supervised FER under more realistic data settings, including OOD and unconstrained unlabeled data. Furthermore, we also report the performance of ID SSL with a small unlabelled set.
We also expand the scope of our study by including a few of the more recent semi-supervised methods. More specifically, we study 11 recent semi-supervised approaches, namely Pi-model \cite{pi_model}, Pseudo-label \cite{pseudo_labels}, Mean Teacher \cite{mean_teacher}, VAT \cite{vat}, UDA \cite{uda}, MixMatch \cite{mixmatch}, ReMixMatch \cite{remixmatch}, FlexMatch \cite{flexmatch}, CoMatch \cite{comatch}, and CCSSL \cite{ccssl}. 
We conduct extensive experiments for different unlabelled set configurations using five public FER datasets, namely FER13 \cite{fer13}, RAF-DB \cite{raf_db}, AffectNet \cite{affectnet}, KDEF \cite{kdef}, and DDCF \cite{ddcf}, along with a non-FER dataset for the unconstrained data, namely CelebA \cite{celeba}. Figure \ref{fig:bannar} depicts an overview of our study on semi-supervised learning from different unlabelled set configurations. Findings from these experiments suggest that FixMatch performs the best among the 11 semi-supervised methods for conventional ID semi-supervised learning, but it performs poorly in other challenging settings. For both OOD and unconstrained unlabelled data, ReMixMatch exhibits the best performance. ReMixMatch also outperforms other methods in the low-data scenario. We also show hyper-parameter sensitivity studies for each of these semi-supervised settings to further boost performance. 

Our contributions in this work are four-fold:
\begin{itemize}
    \item We present a comprehensive and extensive study on 11 recent semi-supervised methods for FER and their comparison to fully supervised learning using six public datasets. 
    \item We compare the performance of all methods under various data scenarios where the unlabelled data is ID, OOD, unconstrained, or very small. 
    \item Our study finds that FixMatch consistently exhibits the best performance for ID unlabelled data, while ReMixMatch is the top-performing approach for OOD, unconstrained, and scarce unlabelled data. We also find that semi-supervised learning improves performance over supervised learning in all the tested scenarios.
    \item We make our code publicly available for quick reproducibility and further developments in this field: \href{https://github.com/ShuvenduRoy/SSL\_FER\_OOD}{github.com/ShuvenduRoy/SSL\_FER\_OOD}. 
\end{itemize}

\begin{figure*}[t]
\centering
\includegraphics[width=0.8\textwidth]{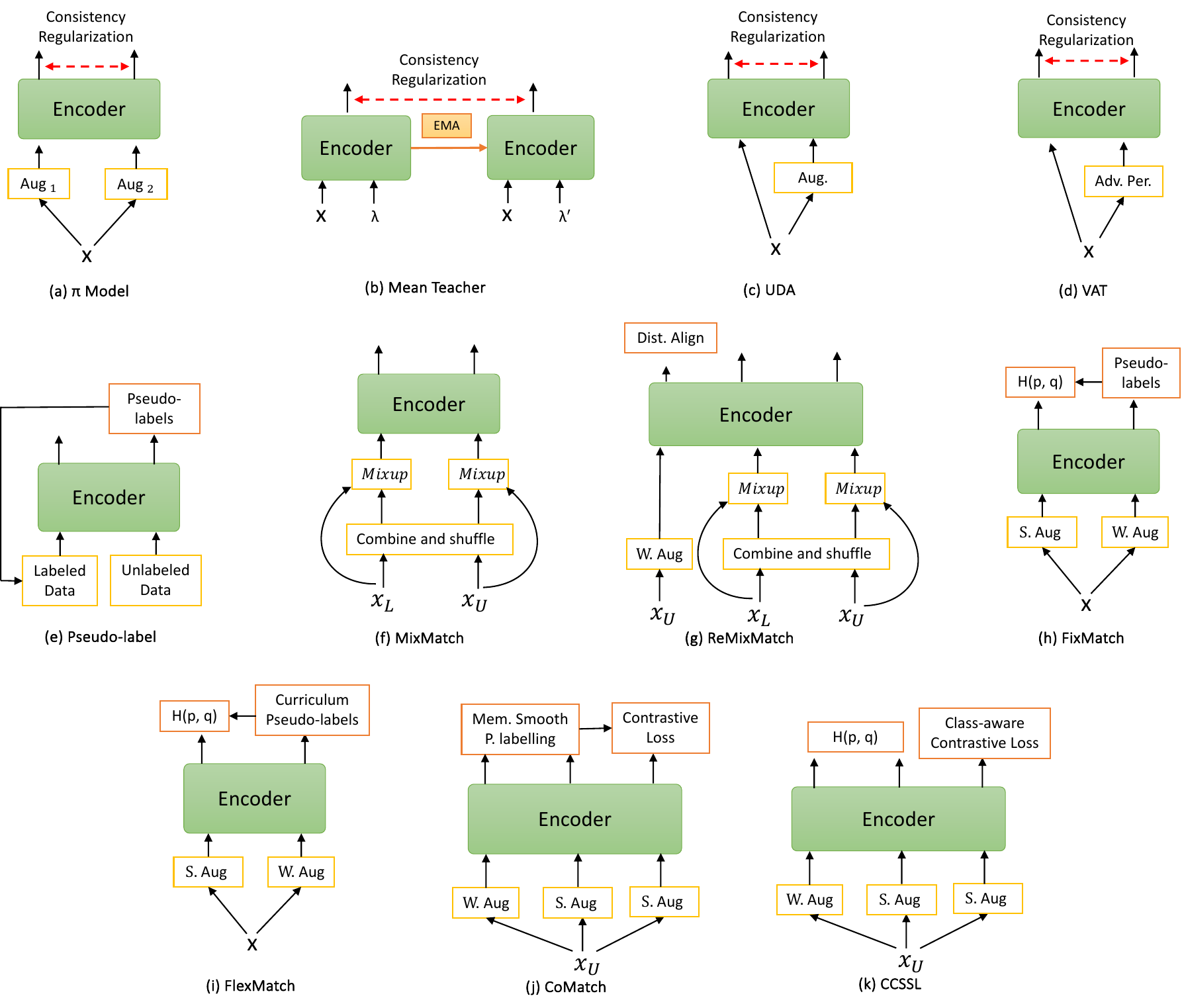}
\caption{Overview of the semi-supervised learning methods explored in this study. \textcolor{black}{Here, $\textit{\text{Aug}}_\textit{\text{i}}$, \textit{S. Aug}, \textit{W. Aug} and \textit{MixUp} refer to the \textit{i}th augmentation of the input $x$, a strongly augmented image, a weekly augmented image, and an augmented image with \textit{MixUp} operation. Consistency Regularization is different across methods, as defined in Eqs. \ref{eq_pimodel}, \ref{eq_meanteacher}, \ref{eq_uda}, and \ref{eq_vat} for Pi-Model, Mean Teacher, UDA, and VAT, respectively. \textit{EMA} refers to the exponential moving average. Adv. Per. refers to adversarial perturbation. $H(p, q)$ is the cross-entropy loss. \textit{Dist. Align} is the distribution alignment concept introduced in ReMixMatch. Curriculum pseudo-labels are generated by the concept of adaptive threshold in FlexMatch. \textit{Mem. Smooth P. labelling} is the concept of memory-smoothed pseudo-labels introduced in CoMatch.}}
\label{fig:models}
\end{figure*}

\section{Related Work}\label{sec:related_work}

In this section, we review the existing literature on semi-supervised learning from two main perspectives that are relevant to our work: (a) with ID data, (b) with OOD data, and (c) semi-supervised methods used specifically for FER.

\subsection{ID SSL}
In recent years, there has been an increasing interest in the literature on semi-supervised learning due to its potential for training large models with small amounts of labelled data. Most of these methods have been demonstrated to perform well on unlabelled data that come from the same distribution as the labelled set. The existing literature on semi-supervised learning can broadly be divided into two categories: entropy minimization \cite{pseudo_labels,noisystudent} and consistency regularization \cite{mean_teacher,uda,vat}.
 
Entropy minimization-based methods learn by predicting the pseudo-labels of unlabelled samples with low entropy. Pseudo-label \cite{pseudo_labels} was one of the first works in this direction, where the pseudo-labels for unlabelled samples are predicted and added to the labelled set if their entropy is low. Noisy-Student \cite{noisystudent} generates these pseudo-labels with a pre-trained encoder, while Meta Pseudo-label \cite{MPL} uses a teacher network to update the pseudo-labels based on the student network's performance. In contrast, consistency regularization-based methods aim to generate consistent predictions for different perturbations of the same input. Pi-model \cite{pi_model} was one of the first works in this direction, where two augmentations of an unlabelled image are forced to generate the same class prediction. Virtual adversarial training (VAT) \cite{vat} replaces the augmentation with adversarial perturbations and enforces consistency on the predictions. Unsupervised domain adaptation (UDA) \cite{uda} shows that replacing simple augmentations with hard augmentations \cite{randaugment,augmix} could bring significant improvement to semi-supervised methods. Since then, most of the semi-supervised methods have used some form of hard augmentation in their pipeline.

Another line of work combines these two prominent approaches into the same framework. For example, MixMatch \cite{mixmatch} enforces consistency on two perturbations (generated with MixUp \cite{mixup}) and optimizes for lower entropy in the predictions. ReMixMatch \cite{remixmatch} improves upon MixMatch by introducing two new concepts: augmentation anchoring and distribution alignment. Another hybrid method is FixMatch \cite{fixmatch}, which has gained tremendous success because of its simplicity while achieving state-of-the-art results in various domains. FixMatch learns by predicting the pseudo-labels for unlabelled images from its weak augmentation and uses them as ground truth for the hard augmentation of the same image if the confidence of the prediction is higher than a threshold. Several improvements have been made to FixMatch since its introduction, such as FlexMatch \cite{flexmatch}, which introduces the curriculum concept to adjust the threshold for different classes dynamically, and CoMatch \cite{comatch}, which introduces an extra contrastive loss term guided by the predicted pseudo-labels.

\textcolor{black}{
\subsection{OOD and Unconstrained SSL}
Since collecting a large amount of ID unlabelled data is difficult in practice, some recent semi-supervised methods have focused on learning from OOD \cite{yoshihashi2019classification, guo2020safe} or unconstrained unlabelled data \cite{auxmix,UnMixMatch}. In OOD SSL \cite{yoshihashi2019classification, guo2020safe, ccssl}, samples in the unlabelled set belong to the same classes as the labelled set but come from a different source and, therefore, have a different data distribution. In some of the earlier works \cite{yoshihashi2019classification}, the main idea was to incorporate an OOD detection module to identify and remove OOD samples from the unlabelled set, effectively focusing on learning from the ID unlabelled data. 
Finally, unconstrained SSL assumes that the unlabeled data are out-of-distribution relative to the labelled set. Additionally, this data is not necessarily limited to the classes present in the labelled set.
CCSSL \cite{ccssl} proposed a method that uses class information from the labelled set along with contrastive learning to effectively learn from unconstrained unlabelled data. 
AuxMix \cite{auxmix} combined self-supervised learning with a novel entropy maximization technique to learn the representations from the unconstrained unlabelled data. 
UnMixMatch \cite{UnMixMatch} employed hard augmentation (RandAugMix) for learning from labelled data, coupling it with contrastive learning and a rotation prediction task for learning from unconstrained unlabelled data.
}

\subsection{Semi-supervised FER} 
Besides our previous work, which focused on benchmarking the most commonly used semi-supervised methods for FER \cite{acii2022}, there are few other studies in this area. For instance, \cite{kurup2019semi} investigated the use of Deep Belief Networks for semi-supervised FER and found that they produced a relative improvement over supervised methods. In \cite{cohen2003semi}, a Bayesian network was proposed for semi-supervised FER. More recently, \cite{li2022towards} proposed an entropy-minimization method for semi-supervised FER, which introduced an adaptive confidence margin concept to partition the unlabelled data based on the confidence of pseudo-labels. The method was then trained on low- and high-confidence predictions separately. Furthermore, \cite{li2022towards} explored the use of multi-modal data to learn from audio-visual signals in a semi-supervised setting. 
\textcolor{black}{Progressive Teacher (PT) \cite{PT} introduced the concept of identifying and selecting useful samples 
for supervised learning, along with a consistency loss on the unlabelled data. To address class distribution mismatch between labelled and unlabelled data, Rethink-Self-SSL \cite{Rethink-Self-SSL} introduced a clustering concept that leveraged intra-cluster and inter-cluster distances to accurately identify out-of-distribution data. To mitigate the impact of false pseudo-labels on model performance, CFRN \cite{CFRN} introduced a feature dropout and emphasis module, enhancing its ability to discriminate between features effectively.}

\section{Method}\label{sec:method}
In this section, we first discuss the problem setup for semi-supervised learning. This is followed by an overview of 11 semi-supervised methods explored in this study. 

\subsection{Problem Setup for Semi-supervised Learning}
Let be given a small labelled set $D_l=\{(x_i^l,y_i^l)\}^N_{i=1}$, where $N$ is the total number of samples and their corresponding class labels, and a large unlabelled set $D_u=\{(x_i^u)\}^M_{i=1}$, where $M$ is the total number of unlabelled samples and $M\gg N$. Accordingly, semi-supervised learning aims to learn from labelled set $D_l$ in a supervised setting while utilizing the unlabelled set $D_u$ in an unsupervised setting to learn a better representation of the data. The performance is validated on a separate validation set $D_v=\{(x_i,y_i)\}^V_{i=1}$. There is no overlap between the sample in labelled, unlabelled, and validation sets, i.e., $D_l \cap D_u \cap D_v = \emptyset $.

\subsection{Semi-Supervised Methods}

\subsubsection{Pi-Model}
Pi-Model \cite{pi_model} is one of the earliest and most popular consistency regularization-based semi-supervised methods. While learning from the labelled samples in a supervised setting, Pi-model learns from the unlabelled set with a consistency regularization term. More specifically, it applies two augmentations on an unlabelled image and forces their predictions to be similar. Pi-Model also uses dropout and random max-pooling to add stochastic behaviour to the predictions. A schematic diagram of the Pi-Model is presented in Figure \ref{fig:models}a. The consistency regularization loss of Pi-Model is represented as
\begin{equation}\label{eq_pimodel}
		\mathbb{E}_{x\in D_l} \mathcal{R}(f(\theta, \tau_1(x)),f(\theta, \tau_2(x))),
\end{equation}
where $\tau_1$ and $\tau_2$ are two sample transformations applied to an unlabelled sample $x$, $f$ is the model, and $\theta$ represents the parameters of $f$.

\subsubsection{Mean Teacher}

Mean Teacher \cite{mean_teacher} is another consistency regularization-based semi-supervised method that is built upon Pi-model. However, Mean-teacher differs in the way it generates embeddings of two augmented samples. Instead of utilizing the online encoder (trainable encoder) to make the prediction for both images, Mean-teacher uses an exponential moving average (EMA) encoder to generate a prediction for the second image. This EMA of the student model is referred to as the teacher model, while the online encoder is called the student model. So, the Mean-teacher learns from unlabelled data by enforcing consistency between the predictions of the teacher and student models on two augmentations of the same sample through a regularization loss. A diagram demonstrating the Mean teacher method is presented in Figure \ref{fig:models}b. The consistency regularization loss for Mean Teacher can be expressed as:
\begin{equation}\label{eq_meanteacher}
		\mathbb{E}_{x\in D_u} \mathcal{R}(f(\theta, \tau_1(x)),f(EMA(\theta), \tau_2(x))),
\end{equation}
where $EMA(\theta)$ is the parameters of the teacher model. Finally, the weight update operation of the teacher model with an exponential moving average is formulated as:
\begin{equation}
    \theta'= m \theta'+ (1-m)\theta,
\end{equation}
where $m$ is a momentum coefficient.

\begin{figure}[!t]
\centering
\includegraphics[width=0.85\linewidth]{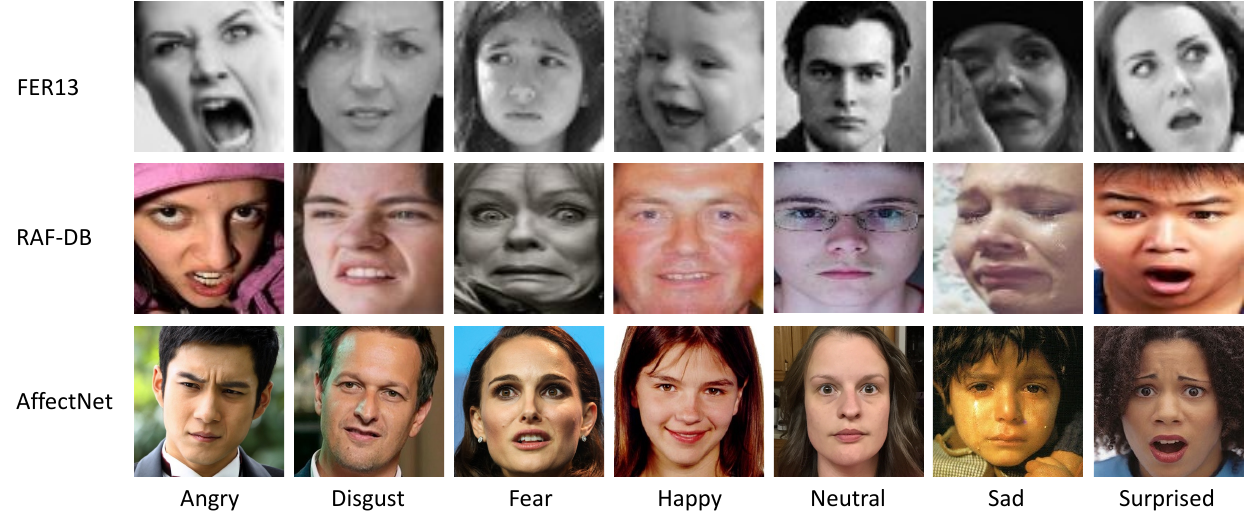}
\caption{Sample images from three main FER datasets: FER13, RAF-DB, and AffectNet.}
\label{fig:example}
\end{figure}

\subsubsection{UDA}
UDA \cite{uda} is also based on the consistency regularization concept. The basic idea of UDA is similar to Pi-model but shows a large improvement in performance, only replacing the augmentation module. UDA replaces the usual augmentation module with advanced augmentation methods like AutoAugment \cite{autoaugment}, and RandAugment \cite{randaugment}, resulting in dynamic and diverse sample augmentations.
Figure \ref{fig:models}c shows an overview of the UDA method. The consistency regularization loss of UDA can be expressed as:
\begin{equation}\label{eq_uda}
		\mathbb{E}_{x\in D_u} \mathcal{R}(f(\theta, x),f(\theta, \tau(x))),
\end{equation}
where $\tau$ represents hard augmentations.

\subsubsection{VAT}
VAT \cite{vat} is similar to the Pi-model and UDA in terms of its regularization concept. However, instead of regularizing the embeddings of two augmented versions of the same sample, VAT uses adversarial perturbation as a different form of augmentation of the input sample. The overview of VAT is depicted in Figure \ref{fig:models}d. The consistency regularization loss of VAT can be expressed as follows: 
\begin{equation}\label{eq_vat}
		\mathbb{E}_{x\in D_u} \mathcal{R}(f(\theta, x),f(\theta, \gamma^{adv}(x))),
\end{equation}
where $\gamma^{adv}$ is the adversarial perturbation operation.

\subsubsection{Pseudo-label}
The Pseudo-label method, introduced in \cite{pseudo_labels}, is an entropy minimization-based method that presents a simple yet effective semi-supervised solution. Pseudo-label involves predicting the class probabilities for each of the unlabelled samples, which act as pseudo-labels for those images. If the confidence of the prediction is high (low entropy), the pseudo-label of the unlabelled sample is treated as a label to train alongside labelled data. This pseudo-labelling concept has been used as a basis for several of the current state-of-the-art methods. A visual illustration of the Pseudo-label method is depicted in Figure \ref{fig:models}e. The loss function of the Pseudo-label method is expressed as:
\begin{equation}
		\mathcal{L}=\mathcal{L}(y_i^l, f(\theta, x_i^l))+\lambda  \mathcal{L}(y_i^{u},f(\theta, x_i^u)),
	\end{equation}
where $y_i^{u}$ is the predicted pseudo-label for an unlabelled sample $x_i^{u}$, and $\lambda$ is a coefficient to balance the weight of two loss terms.

\subsubsection{MixMatch}
MixMatch \cite{mixmatch} is a semi-supervised method of the hybrid category that combines the concept of consistency regularization and entropy minimization. Similar to the entropy-based methods, MixMatch aims to generate low entropy predictions on the unlabelled data and also enforces consistency in its predictions similar to consistency-based methods. The novel component of MixMatch is the MixUp operation (an interpolation function) on labelled and unlabelled samples to generate mixed samples. Both entropy minimization and consistency regularization operations are applied to the mixed samples. The MixUp operation can be represented as: 
\begin{equation}
    x'=\alpha x_l + (1-\alpha)x_u,
\end{equation}
where $x_l$ and $x_u$ are input samples from the labelled and unlabelled set, and $\alpha$ is the weight factor that balances the labelled and unlabelled components in the generated sample. MixMatch uses a beta distribution to randomly sample the value for alpha.  

Another key innovation of MixMatch is the use of multiple augmentations on the unlabelled set and averaging the predictions on the augmented samples to generate the final prediction for that unlabelled sample. Let $d_l^\prime$ and $d_u^\prime$ be a batch of labelled and unlabelled samples after the MixUp operation. The MixMatch loss can be represented as: 
\begin{align}
    \mathcal{L}_l &= \frac{1}{|d_l^\prime|} \sum_{x, y \in d_l^\prime} H(y, f(x, \theta)) \label{eqn:l_x} , \\
    \mathcal{L}_u &= \frac{1}{C|d_u^\prime|} \sum_{x^\prime, y^\prime \in d_u^\prime} \|y\prime - f(x, \theta)\|_2^2 \label{eqn:l_u} , \\
    \mathcal{L} &= \mathcal{L}_l + \lambda \mathcal{L}_u . \label{eqn:l_combined}
\end{align}
where $H(.)$ is the cross-entropy loss, $C$ is the total number of classes, and $\lambda$ is the weight factor between the supervised and unsupervised loss terms. Figure \ref{fig:models}f shows a visual illustration of the MixMatch method.  

\subsubsection{ReMixMatch} 

ReMixMatch \cite{remixmatch} is a modified version of MixMatch that incorporates two new concepts: distribution alignment and augmentation anchoring. The former aims to ensure that the predictions made on the unlabelled data align with the distribution of the predictions on the labelled data. Augmentation anchoring replaces the consistency regularization of MixMatch and focuses on making the representation of strongly augmented samples similar to those of weakly augmented samples. This technique compares one weakly augmented sample against multiple strongly augmented samples. ReMixMatch also introduces a new strong augmentation method called CTAugment, which is more suitable in semi-supervised learning settings. Figure \ref{fig:models}h provides a visual representation of the ReMixMatch.

\subsubsection{FixMatch}

FixMatch \cite{fixmatch} is another hybrid semi-supervised learning method that shows impressive performance in many applications. For an unlabelled sample, FixMatch first applies weak augmentations and generates a prediction. FixMatch then considers this as a pseudo-label for a hard augmentation of the same sample if the confidence of this pseudo-label is beyond a threshold. Standard shift and flip augmentations are utilized for the weak augmentation module of FixMatch. FixMatch explores RandAugment \cite{randaugment} and CTAugment \cite{remixmatch} as the hard augmentation module. The unsupervised loss of FixMatch can be represented as:
\begin{equation}
    \mathcal{L}_u = \frac{1}{|d_u|} \sum_{x\in d_u} \mathbf{1}(max(q) \geq \tau)~  \mathcal{H}(\hat{q}, f(\mathcal{A}(x), \theta)) \label{eqn:fixmatch}
\end{equation}
where, $\tau$ is the threshold, $q$ is the prediction on the weakly augmented sample, and $\hat{q} = arg~max (q)$. A visual illustration of FixMatch is depicted in Figure \ref{fig:models}h.

\subsubsection{FlexMatch}
FlexMatch \cite{flexmatch} proposes an improvement over FixMatch with a curriculum learning concept for the threshold parameter. Rather than using a fixed $\tau$ for all classes, FlexMatch updates a class-specific threshold based on the learning status of that class. FlexMatch uses per-class accuracy as an indicator of the learning status of that class, which is calculated as:
\begin{equation}
    \alpha(c) = \sum_{n} \mathbf{1}(max(q) \geq \tau)~  \mathbf{1}(argmax(q))= c), \label{eqn:fixmatch}
\end{equation}
where $n$ is the number of samples. A schematic diagram of FlexMatch is shown in Figure \ref{fig:models}i.

\subsubsection{CoMatch}
\textcolor{black}{
CoMatch \cite{comatch} introduces a graph contrastive learning concept built on Fixmatch. CoMatch jointly learns two representations of data that interact and improve with each other: class probabilities and low-dimensional embeddings. To reduce the errors in the predicted pseudo-labels, CoMatch uses the concept of memory-smoothed pseudo-labels, where label predictions are refined by considering similar data points in the embedding space. To learn better task-specific representations, CoMatch uses contrastive learning to encourage similar embeddings for samples with the same label. The Concept of CoMatch is depicted in Figure \ref{fig:models}j.}

\subsubsection{CCSSL}
\textcolor{black}{
Finally, CCSSL \cite{ccssl} is also built on FixMatch, and deals with the confirmation bias of pseudo-labels to improve performance on OOD unlabelled data. Unlike traditional pseudo-labelling approaches, CCSSL separates data into in-distribution and out-of-distribution categories. It then applies class-wise clustering to maintain efficient learning for known categories in the in-distribution data, while employing image-level contrastive learning on out-of-distribution data. Overall, CCSSL is designed as an add-on that can be easily integrated with existing pseudo-labeling methods, enhancing their effectiveness and making them more applicable in diverse real-world scenarios. The CCSSL method is illustrated in Figure \ref{fig:models}k.
}

\section{Experiments}\label{sec:results}
In this section, 
we first provide details of the datasets used in this study. Following this, we present the experimental setup employed for the semi-supervised FER experiments. Finally, we present the main results for all the semi-supervised settings.

\begin{figure}
    \centering

     \begin{subfigure}[b]{0.10\textwidth}
         \centering
         \includegraphics[width=.9\textwidth]{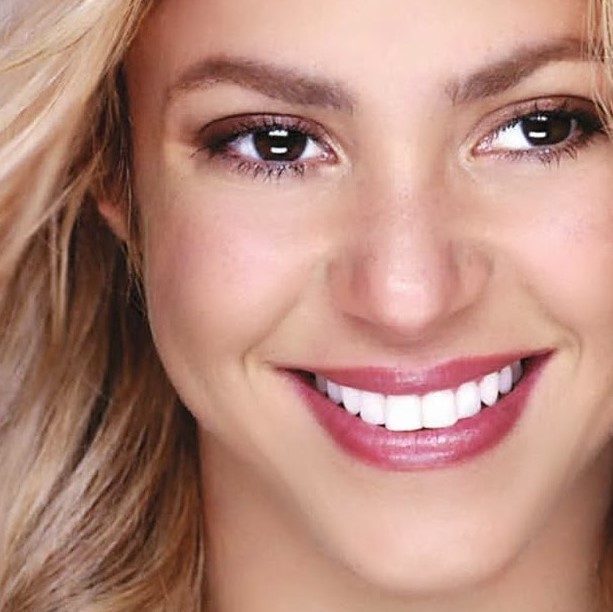}
         \caption{\scriptsize Crop}
     \end{subfigure}
     \begin{subfigure}[b]{0.10\textwidth}
         \centering
         \includegraphics[width=.9\textwidth]{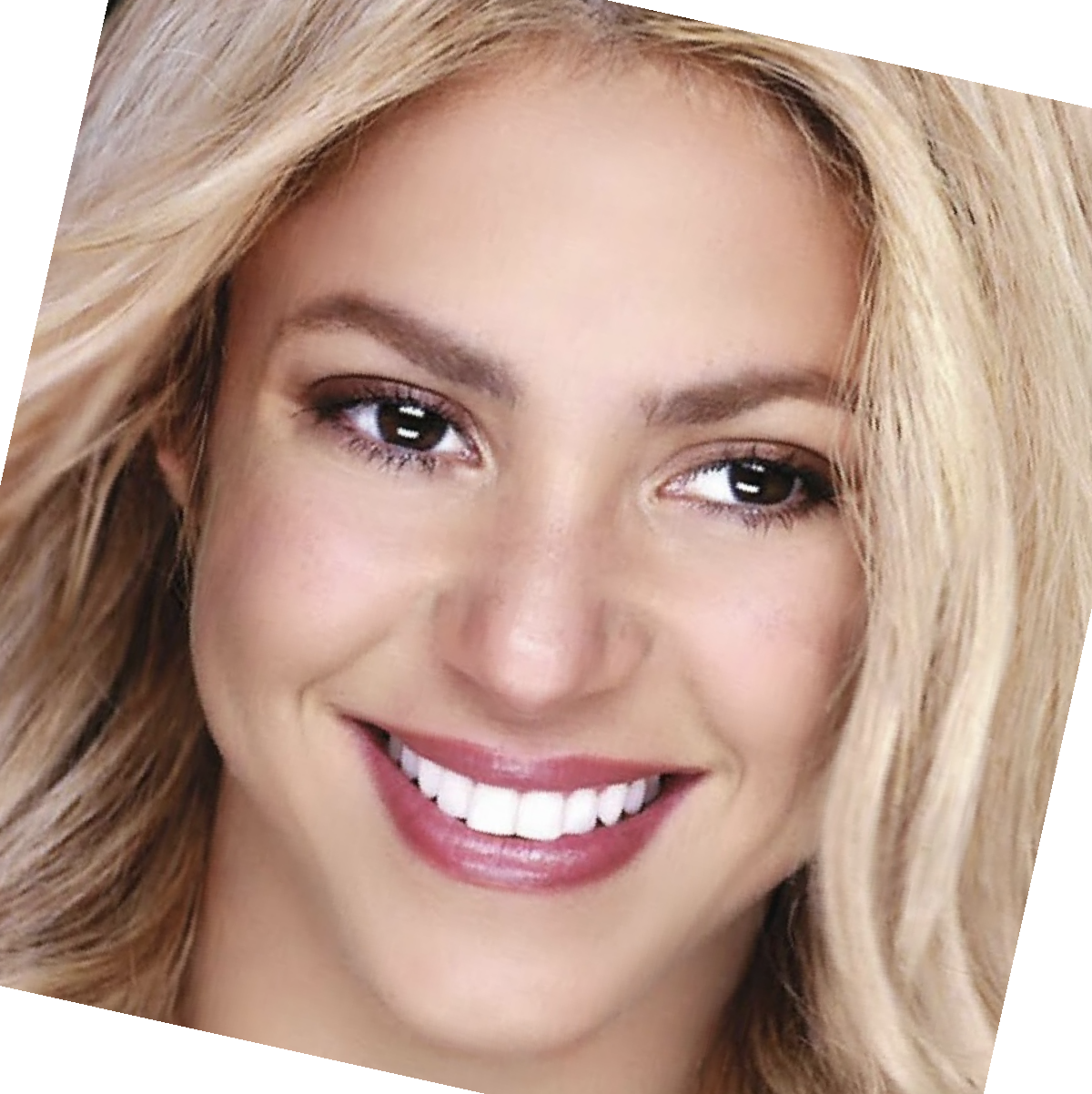}
         \caption{\scriptsize Rotation}
         \label{fig:lambda_mixmatch}
     \end{subfigure}
     \begin{subfigure}[b]{0.10\textwidth}
         \centering
         \includegraphics[width=.9\textwidth]{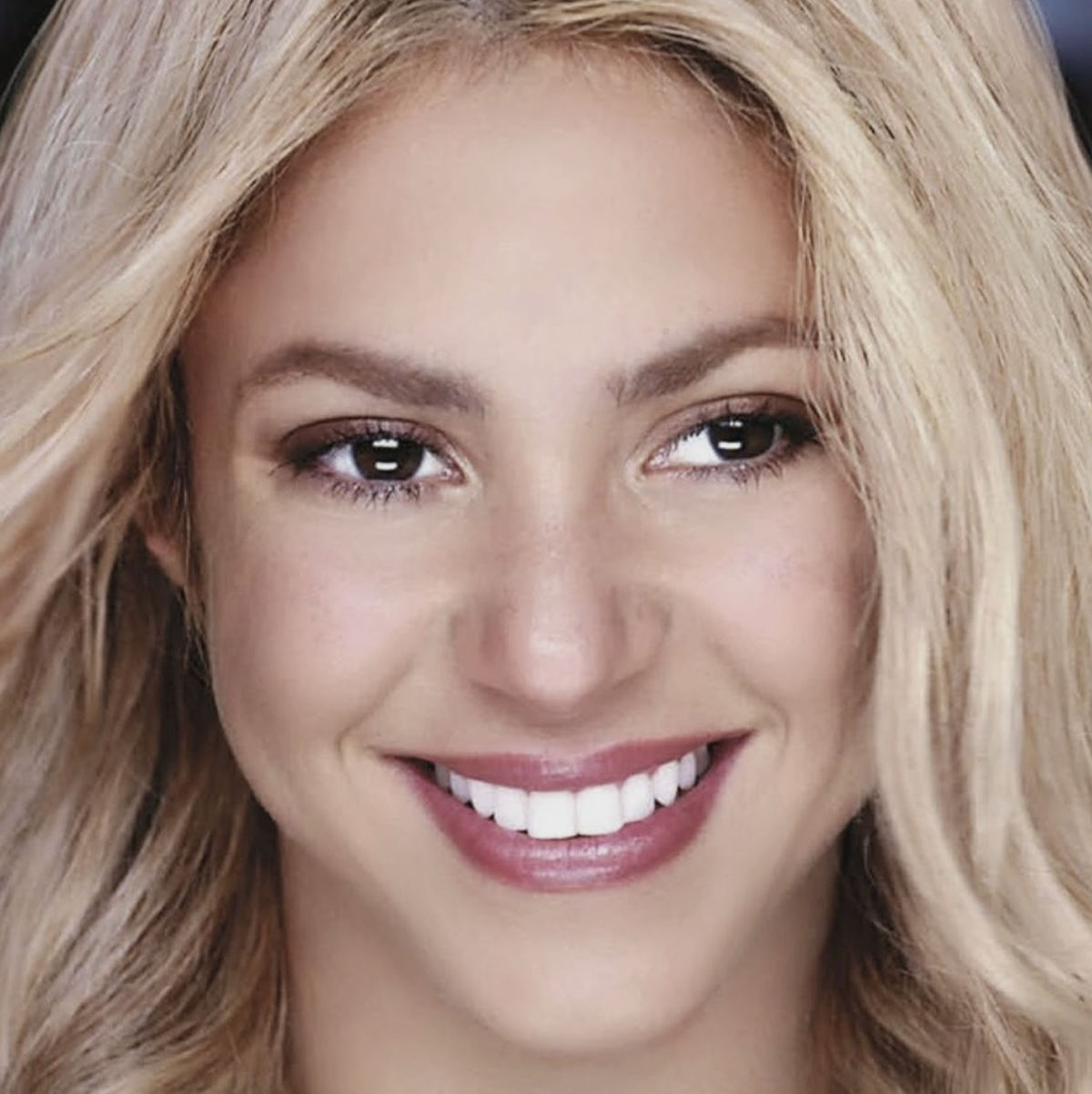}
         \caption{\scriptsize Color Jitter}
         \label{fig:cutoff_fixmatch}
     \end{subfigure}
     \begin{subfigure}[b]{0.10\textwidth}
         \centering
         \includegraphics[width=.9\textwidth]{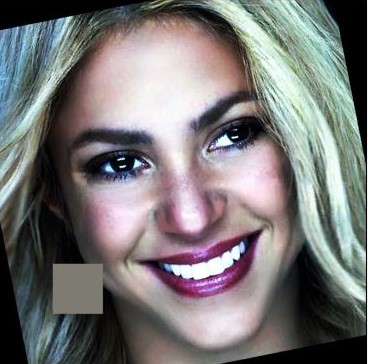}
         \caption{\scriptsize Multiple}
         \label{fig:cutoff_fixmatch}
     \end{subfigure}
\caption{Examples of hard augmentations.}
\label{fig:aug}

\end{figure}

\subsection{Datasets} 

In this study, we utilize a total of six datasets to conduct the Facial Expression Recognition (FER) experiments. For the main results, we use three datasets: FER13 \cite{fer13}, RAF-DB \cite{rafdb}, and AffectNet \cite{affectnet}. In addition, for the experiments on unconstrained FER, we use the CelebA \cite{celeba} dataset. Finally, for the FER experiments on limited data, we use the KDEF \cite{kdef} and DDCF \cite{ddcf} datasets. Below, we provide a brief description of each dataset used in this study. Some samples from these datasets are shown in Figure \ref{fig:example}.

\textbf{FER13} \cite{fer13} is a widely used dataset for FER that contains over 28.7K images of 7 emotions (anger, disgust, fear, happiness, sadness, surprise, and neutral). The images have been collected from the internet and resized to 48$\times$48 pixels.

\textbf{RAF-DB} \cite{rafdb} is a dataset for FER that contains around 15K images, with 12K images for training and 3K used for testing. This dataset has been annotated by 315 annotators, with each image annotated by around 40 annotators.

\textbf{AffectNet} \cite{affectnet} is a large-scale dataset for FER that contains around 284K images of 8 emotions (anger, contempt, disgust, fear, happiness, sadness, surprise, and neutral). This dataset has also been collected from the internet, and the images are of relatively high resolution.

\textbf{CelebA} \cite{celeba} is a large-scale dataset of faces with aourd 202K images. This dataset does not necessarily contain images of expressive faces, and the images are also collected from the internet and are of relatively high resolution.

\textbf{KDEF} \cite{kdef} is a small dataset for FER that contains images of 7 emotions, which have been captured in lab conditions. This dataset contains a total of approximately 5K images. The images are of relatively high resolution and have been captured from different angles.

\textbf{DDCF} \cite{ddcf} is another small dataset for FER that contains images of 8 emotions, which have been captured in lab conditions. There are nearly 6.5K images in this dataset. The images are of relatively high resolution and have been collected from different angles.

\subsection{Implementation details}
For a fair comparison between all the methods, we use the same encoder and training protocol.
For the encoder, we use ResNet-50 \cite{resnet}. 
All the experiments are reported for different numbers of labelled samples. For this, we randomly sample $N=n\times C$ images and their corresponding labels from the labelled set $D_l$, where $C$ is the number of classes, and $n$ is the number of samples per class. Specifically, we present all results for $n \in \{10, 25, 100, 250 \}$. \textcolor{black}{For all experiments, we report the average accuracy and standard deviations over \textit{three} runs on different seeds. Following \cite{fixmatch}, we sample different random splits of data with different seeds.}

We follow the implementation details of the original methods for method-specific parameters. For instance, we use a confidence cut-off value of 0.95 for FixMatch, a moving average weight of 0.999 for Mean-teacher, and a temperature value of 0.5 for methods that utilize sharpening distribution. For training, we use $2^{20}$ iterations with a batch size of 64 and SGD optimizer with a learning rate of 0.03, a momentum of 0.9, and a cosine learning rate decay scheduler.

In many recent semi-supervised methods, data augmentations are proven to be an essential component. 
\textcolor{black}{In this context, a hard augmentation refers to a sequence of augmentations applied to a sample that results in an augmented sample that is visually distinguishable from the original input. In contrast, weak augmentation refers to a single or very low number of augmentations ($\leq$ 2) applied to an input sample which does not change the sample drastically.}
Among the various hard augmentation modules mentioned in the literature, RandAugment \cite{randaugment} is the most commonly used one. This technique involves defining a sequence (up to 14) of augmentations that can be applied to an image, such as random crops, flips, and colour distortion. Then, a random subset of these augmentations is selected and applied to the image. We provide some examples of the augmentations used in RandAugment in Figure \ref{fig:aug}.

\begin{table*}[ht]
    \centering
    \setlength
    \tabcolsep{2.6pt}
    \caption{The performance of different semi-supervised methods with ID unlabelled data on FER13, RAF-DB, and AffectNet, when 10, 25, 100, and 250 labelled samples per class are used for training. }
    \resizebox{1\textwidth}{!}
    {
    \begin{tabular}{l|rrrr|rrrr|rrrr| c}
    \toprule
    & \multicolumn{4}{c|}{\textbf{FER13}} & \multicolumn{4}{c|}{\textbf{RAF-DB}}  & \multicolumn{4}{c|}{\textbf{AffectNet}} & Avg. Acc \\
    \cmidrule(l{3pt}r{3pt}){1-14}
    Method / $m$                & 10 labels        & 25 labels      & 100 labels     & 250 labels    & 10 labels         & 25 labels      & 100 labels     & 250 labels    & 10 labels         & 25 labels      & 100 labels     & 250 labels      \\
   \cmidrule(l{3pt}r{3pt}){1-14}

    $\Pi$-model \cite{pi_model}	&
    37.09\textcolor{black}{\tiny $\pm 3.7$} &
    40.87\textcolor{black}{\tiny $\pm 2.5$} &
    50.66\textcolor{black}{\tiny $\pm 1.8$} &
    56.42\textcolor{black}{\tiny $\pm 1.4$} &

    39.86\textcolor{black}{\tiny $\pm 3.1$} &
    50.97\textcolor{black}{\tiny $\pm 2.5$} &
    63.98\textcolor{black}{\tiny $\pm 1.1$} &
    71.15\textcolor{black}{\tiny $\pm 0.8$} &

    24.17\textcolor{black}{\tiny $\pm 4.2$} &
    25.37\textcolor{black}{\tiny $\pm 3.8$} &
    31.24\textcolor{black}{\tiny $\pm 3.4$} &
    32.40\textcolor{black}{\tiny $\pm 2.1$} &
    43.68\\

    Pseudo-label \cite{pseudo_labels} &
    32.79\textcolor{black}{\tiny $\pm 3.9$} &
    36.04\textcolor{black}{\tiny $\pm 2.7$} &
    49.21\textcolor{black}{\tiny $\pm 1.9$} &
    54.88\textcolor{black}{\tiny $\pm 1.5$} &

    58.31\textcolor{black}{\tiny $\pm 3.5$} &
    39.11\textcolor{black}{\tiny $\pm 2.6$} &
    54.07\textcolor{black}{\tiny $\pm 1.7$} &
    67.40\textcolor{black}{\tiny $\pm 0.9$} &

    18.00\textcolor{black}{\tiny $\pm 4.4$} &
    21.05\textcolor{black}{\tiny $\pm 3.0$} &
    33.05\textcolor{black}{\tiny $\pm 3.6$} &
    37.37\textcolor{black}{\tiny $\pm 2.3$} &
    41.77\\

    Mean Teacher \cite{mean_teacher} &
    45.21\textcolor{black}{\tiny $\pm 2.6$} &
    \underline{55.14}\textcolor{black}{\tiny $\pm 1.8$} &
    52.17\textcolor{black}{\tiny $\pm 1.6$} &
    58.06\textcolor{black}{\tiny $\pm 1.3$} &

    62.05\textcolor{black}{\tiny $\pm 2.9$} &
    45.17\textcolor{black}{\tiny $\pm 2.3$} &
    45.57\textcolor{black}{\tiny $\pm 1.8$} &
    \textbf{76.85}\textcolor{black}{\tiny $\pm 0.5$} &

    19.54\textcolor{black}{\tiny $\pm 3.9$} &
    20.21\textcolor{black}{\tiny $\pm 3.1$} &
    20.80\textcolor{black}{\tiny $\pm 2.8$} &
    44.05\textcolor{black}{\tiny $\pm 1.1$} &
    45.40\\

    VAT \cite{vat}	&
    24.95\textcolor{black}{\tiny $\pm 3.8$} &
    \textbf{55.22}\textcolor{black}{\tiny $\pm 2.0$} &
    51.55\textcolor{black}{\tiny $\pm 1.7$} &
    55.64\textcolor{black}{\tiny $\pm 1.4$} &

    \underline{63.10}\textcolor{black}{\tiny $\pm 3.1$} &
    45.82\textcolor{black}{\tiny $\pm 2.4$} &
    62.05\textcolor{black}{\tiny $\pm 1.5$} &
    59.45\textcolor{black}{\tiny $\pm 1.0$} &

    17.68\textcolor{black}{\tiny $\pm 4.3$} &
    \underline{35.02}\textcolor{black}{\tiny $\pm 3.4$} &
    37.68\textcolor{black}{\tiny $\pm 3.0$} &
    37.92\textcolor{black}{\tiny $\pm 2.0$} &
    45.50\\

    UDA \cite{uda}	&
    \underline{46.72}\textcolor{black}{\tiny $\pm 2.7$} &
    49.89\textcolor{black}{\tiny $\pm 1.9$} &
    50.62\textcolor{black}{\tiny $\pm 1.6$} &
    60.68\textcolor{black}{\tiny $\pm 1.2$} &

    46.87\textcolor{black}{\tiny $\pm 3.0$} &
    \textbf{53.15}\textcolor{black}{\tiny $\pm 2.4$} &
    58.86\textcolor{black}{\tiny $\pm 1.6$} &
    60.82\textcolor{black}{\tiny $\pm 1.0$} &

    27.42\textcolor{black}{\tiny $\pm 4.1$} &
    32.16\textcolor{black}{\tiny $\pm 3.2$} &
    37.25\textcolor{black}{\tiny $\pm 2.8$} &
    37.64\textcolor{black}{\tiny $\pm 1.8$} &
    46.84\\

    MixMatch \cite{mixmatch}	&
    45.69\textcolor{black}{\tiny $\pm 2.6$} &
    46.41\textcolor{black}{\tiny $\pm 1.8$} &
    55.73\textcolor{black}{\tiny $\pm 1.5$} &
    58.27\textcolor{black}{\tiny $\pm 1.2$} &

    36.34\textcolor{black}{\tiny $\pm 3.2$} &
    43.12\textcolor{black}{\tiny $\pm 2.5$} &
    64.14\textcolor{black}{\tiny $\pm 1.0$} &
    73.66\textcolor{black}{\tiny $\pm 0.4$} &
    
    \textbf{30.80}\textcolor{black}{\tiny $\pm 3.0$} &
    32.40\textcolor{black}{\tiny $\pm 3.1$} &
    39.77\textcolor{black}{\tiny $\pm 2.7$} &
    \underline{48.31}\textcolor{black}{\tiny $\pm 1.6$} &
    \underline{47.88}\\

    ReMixMatch \cite{remixmatch}	&
    41.07\textcolor{black}{\tiny $\pm 2.8$} &
    43.25\textcolor{black}{\tiny $\pm 1.5$} &
    44.62\textcolor{black}{\tiny $\pm 1.3$} &
    57.49\textcolor{black}{\tiny $\pm 1.0$} &

    37.35\textcolor{black}{\tiny $\pm 3.3$} &
    42.56\textcolor{black}{\tiny $\pm 2.1$} &
    42.86\textcolor{black}{\tiny $\pm 1.5$} &
    61.70\textcolor{black}{\tiny $\pm 0.8$} &

    29.28\textcolor{black}{\tiny $\pm 3.2$} &
    33.54\textcolor{black}{\tiny $\pm 2.5$} &
    \underline{41.60}\textcolor{black}{\tiny $\pm 1.6$} &
    46.51\textcolor{black}{\tiny $\pm 1.4$} &
    43.48\\

    FixMatch \cite{fixmatch}	&
    \textbf{47.88}\textcolor{black}{\tiny $\pm 2.5$} &
    49.90\textcolor{black}{\tiny $\pm 1.7$} &
    \textbf{59.46}\textcolor{black}{\tiny $\pm 1.0$} &
    \textbf{62.20}\textcolor{black}{\tiny $\pm 0.5$} &

    \textbf{63.25}\textcolor{black}{\tiny $\pm 1.0$} &
    52.44\textcolor{black}{\tiny $\pm 2.2$} &
    \underline{64.34}\textcolor{black}{\tiny $\pm 0.9$} &
    \underline{75.51}\textcolor{black}{\tiny $\pm 0.3$} &

    \underline{30.08}\textcolor{black}{\tiny $\pm 1.9$} &
    \textbf{38.31}\textcolor{black}{\tiny $\pm 1.2$} &
    \textbf{46.37}\textcolor{black}{\tiny $\pm 1.0$} &
    \textbf{51.25}\textcolor{black}{\tiny $\pm 0.6$} &
    \textbf{53.41}\\

    FlexMatch \cite{flexmatch}      &
    39.77\textcolor{black}{\tiny $\pm 2.5$} &
    42.88\textcolor{black}{\tiny $\pm 1.7$} &
    51.14\textcolor{black}{\tiny $\pm 1.3$} &
    56.06\textcolor{black}{\tiny $\pm 1.0$} &

    40.51\textcolor{black}{\tiny $\pm 3.0$} &
    42.67\textcolor{black}{\tiny $\pm 2.1$} &
    50.75\textcolor{black}{\tiny $\pm 1.5$} &
    61.70\textcolor{black}{\tiny $\pm 0.8$} &

    17.20\textcolor{black}{\tiny $\pm 4.1$} &
    19.80\textcolor{black}{\tiny $\pm 3.0$} &
    22.34\textcolor{black}{\tiny $\pm 2.6$} &
    29.83\textcolor{black}{\tiny $\pm 1.7$} &
    39.55\\

    CoMatch   \cite{comatch}      &
    40.24\textcolor{black}{\tiny $\pm 2.7$} &
    49.04\textcolor{black}{\tiny $\pm 1.9$} &
    54.97\textcolor{black}{\tiny $\pm 1.6$} &
    59.47\textcolor{black}{\tiny $\pm 1.2$} &

    40.04\textcolor{black}{\tiny $\pm 3.1$} &
    \underline{52.59}\textcolor{black}{\tiny $\pm 2.4$} &
    \textbf{68.05}\textcolor{black}{\tiny $\pm 1.1$} &
    73.46\textcolor{black}{\tiny $\pm 0.6$} &

    21.23\textcolor{black}{\tiny $\pm 4.2$} &
    23.54\textcolor{black}{\tiny $\pm 3.2$} &
    27.45\textcolor{black}{\tiny $\pm 2.8$} &
    30.31\textcolor{black}{\tiny $\pm 1.9$} &
    45.03\\

    CCSSL \cite{ccssl}          &
    40.23\textcolor{black}{\tiny $\pm 2.8$} &
    45.36\textcolor{black}{\tiny $\pm 1.8$} &
    \underline{57.01}\textcolor{black}{\tiny $\pm 1.4$} &
    \underline{61.77}\textcolor{black}{\tiny $\pm 1.0$} &

    50.59\textcolor{black}{\tiny $\pm 2.8$} &
    51.30\textcolor{black}{\tiny $\pm 2.2$} &
    63.79\textcolor{black}{\tiny $\pm 1.4$} &
    74.93\textcolor{black}{\tiny $\pm 0.7$} &

    16.89\textcolor{black}{\tiny $\pm 4.3$} &
    21.34\textcolor{black}{\tiny $\pm 3.1$} &
    24.46\textcolor{black}{\tiny $\pm 2.7$} &
    28.94\textcolor{black}{\tiny $\pm 1.8$} &
    44.71\\

    \bottomrule
    \end{tabular}
    }
    \label{tab:results}
\end{table*}

\subsection{Semi-supervised FER with ID unlabelled data}
\subsubsection{Setup}
This section presents the main result of 11 semi-supervised learning approaches on the FER13, RAF-DB, and AffectNet datasets for ID FER. As previously mentioned, the results are presented for 10, 25, 100, and 250 labelled samples for each emotion class. The remaining samples from each dataset are treated as unlabelled sets.

\subsubsection{Performance}
The main results for ID semi-supervised learning are presented in Table \ref{tab:results}, which also includes the average accuracy across all settings (4 data splits of 3 datasets) for an overall understanding of the performance of each method. In summary, FixMatch appears to be the most successful semi-supervised method for ID data, as it outperforms other methods on 7 out of 12 settings and achieves the second-best result on the other two settings. 
\textcolor{black}{FixMatch achieves an average accuracy of 53.41\% across all settings, with a maximum standard deviation of 2.5\% on FER-13 with 10 labelled samples per class. The second-best method, MixMatch, achieves an average accuracy of 47.88\%, which is a considerable 5.53\% drop in performance compared to FixMatch.} Therefore, we can conclude that FixMatch is the most robust semi-supervised method for FER with ID unlabelled data.

\subsubsection{Sensitivity study}
\begin{figure}[!t]
    \centering

     \begin{subfigure}[b]{0.15\textwidth}
         \centering
         \includegraphics[width=1.1\textwidth]{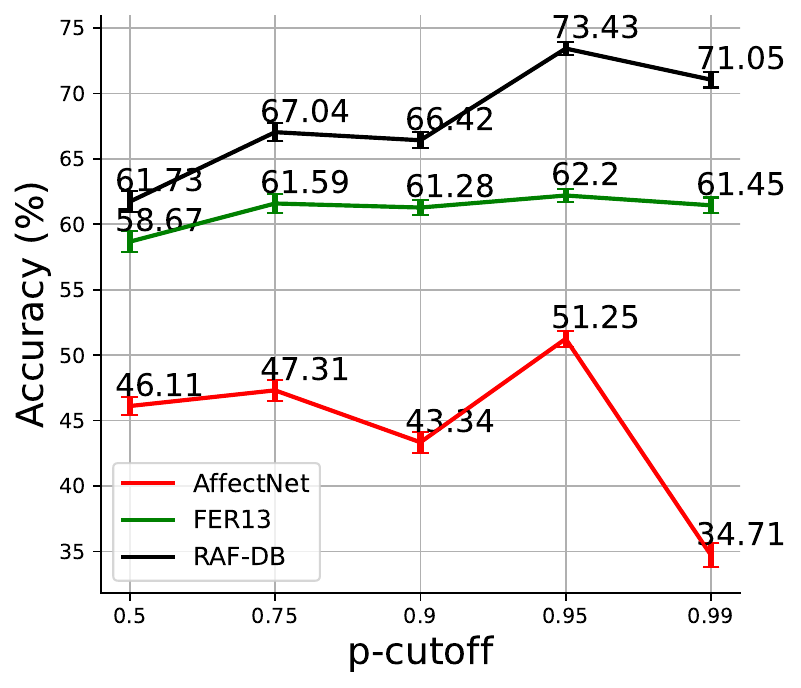}
         \caption{Accuracy vs. $p_{\text{cutoff}}$ for \textit{\textbf{FixMatch}}.}
         \label{fig:cutoff_fixmatch}
     \end{subfigure}
      \hspace{5pt}
      \begin{subfigure}[b]{0.15\textwidth}
         \centering
         \includegraphics[width=1.1\textwidth]{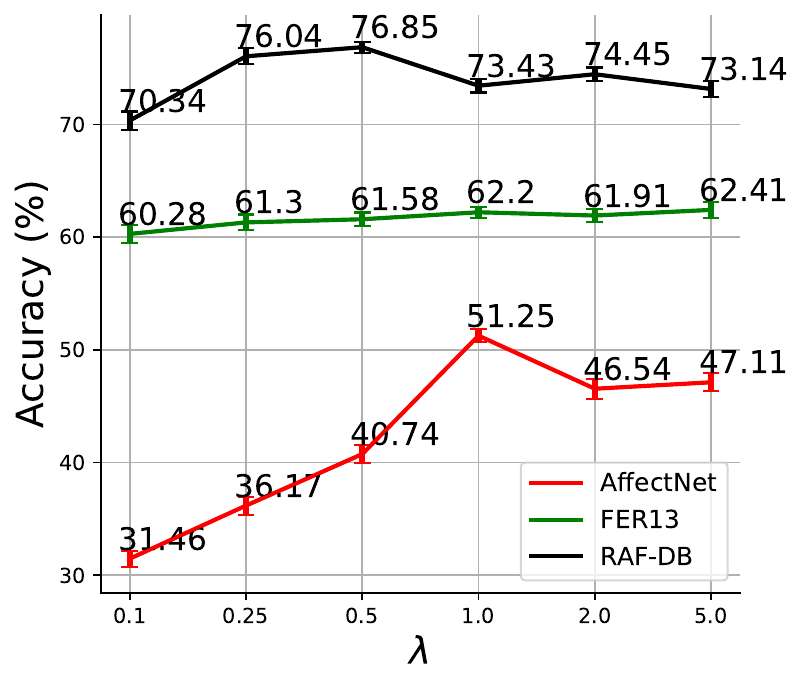}
         \caption{Accuracy vs. $\lambda$ for \textit{\textbf{FixMatch}}.}
         \label{fig:lambda_fixmatch}
     \end{subfigure}
     \\
     \begin{subfigure}[b]{0.15\textwidth}
         \centering
         \includegraphics[width=1.1\textwidth]{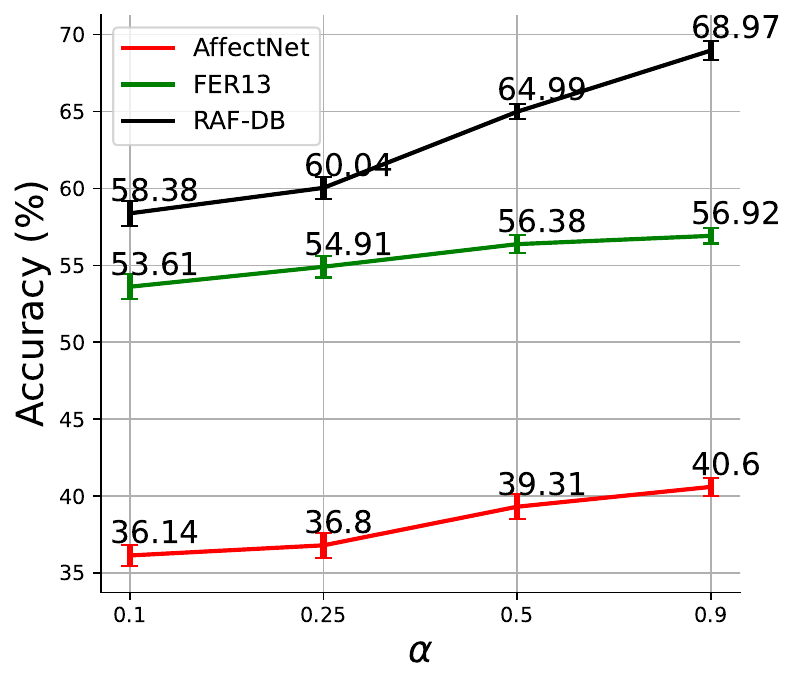}
         \caption{Accuracy vs. $\alpha$ for \textit{\textbf{MixMatch}}.}
         \label{fig:alpha_mixmatch}
     \end{subfigure}
     \hspace{5pt}
      \begin{subfigure}[b]{0.15\textwidth}
         \centering
         \includegraphics[width=1.1\textwidth]{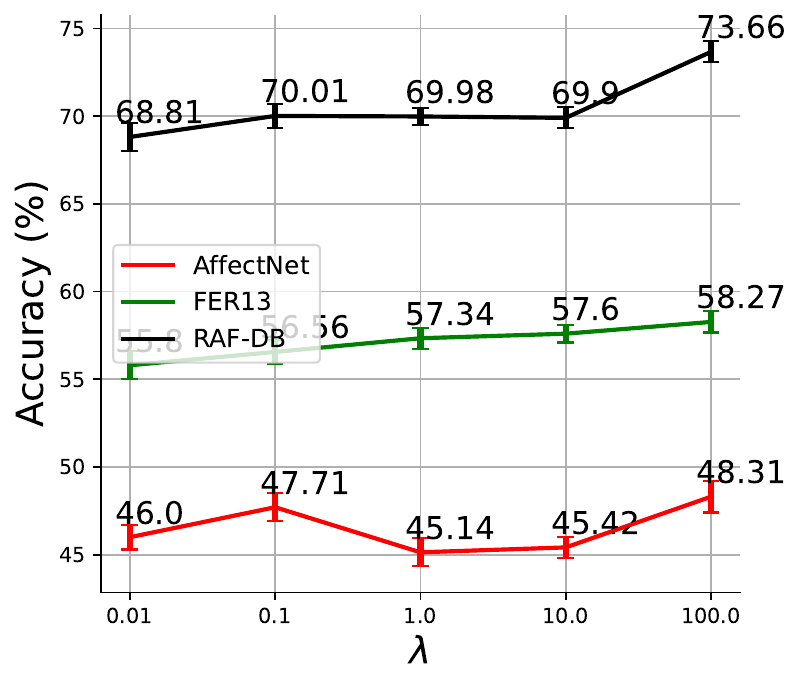}
         \caption{Accuracy vs. $\lambda$ for \textit{\textbf{MixMatch}}.}
         \label{fig:lambda_mixmatch}
     \end{subfigure}

    \caption{Sensitivity study of various parameters for two of the best semi-supervised methods on ID unlabelled data. }
    \label{fig:sensitivity_iid}
\end{figure}

All the experiments shown in the table above were conducted using default parameters for each algorithm, as reported in the original papers. In this subsection, we present a sensitivity study on the key hyper-parameters of the two best-performing methods in order to improve FER performance further. Figure \ref{fig:sensitivity_iid} displays this study on the $P_\text{cutoff}$ and $\lambda$ values for FixMatch, as well as the $\alpha$ and $\lambda$ values for MixMatch. In the FixMatch, the $P_\text{cutoff}$ value determines the confidence threshold at which a predicted pseudo-label is considered as the label for an unlabelled image. The results shown in Figure \ref{fig:cutoff_fixmatch} indicate that the best accuracy is achieved for a $P_\text{cutoff}$ value of 0.95 for all datasets, which is consistent with the original FixMatch method. The $\lambda$ value balances the weight of supervised and unsupervised loss in FixMatch. We conduct an experiment on $\lambda$ (Figure \ref{fig:lambda_fixmatch}) and find that the optimal value varies for different datasets. While AffectNet and RAF-DB show better results for relatively smaller values of $\lambda$ (1.0 and 0.5, respectively), FER13 achieves the best performance with higher values of $\lambda$, specifically 5.0. In the MixMatch method, $\alpha$ is the mixing coefficient used in the MixUp operation. The experiment on $\alpha$ (Figure \ref{fig:alpha_mixmatch}) indicates that higher values of $\alpha$ generally produce better results for all datasets, with the best performance obtained with $\alpha = 0.9$. Finally, in the experiment on the $\lambda$ value of MixMatch (Figure \ref{fig:lambda_mixmatch}), we observe improvements for larger values of $\lambda$, with the best accuracy achieved when $\lambda$ is set to \textit{100} for all three datasets.

\begin{table}[]
\caption{Comparison between fully-supervised learning and semi-supervised learning with ID unlabelled data. }
\begin{center}
\begin{tabular}{l|ll|l l}
\toprule
& \multicolumn{2}{c|}{\textbf{Supervised}} & \multicolumn{2}{c}{\textbf{Semi-sup. (best)}} \\
\cmidrule(l{3pt}r{3pt}){1-5}
\textbf{Dataset}  & \textbf{250/class} & \textbf{All data}  & \textbf{250/class} & \textcolor{black}{\textbf{All data}} \\ 
\cmidrule(l{3pt}r{3pt}){1-5}
FER13 & 
\textcolor{black}{53.58}\textcolor{black}{\tiny $\pm 1.1$} & 
64.57\textcolor{black}{\tiny $\pm 0.9$}&  
\textcolor{black}{62.20}\textcolor{black}{\tiny $\pm 0.5$} & 
\textcolor{black}{65.15}\textcolor{black}{\tiny $\pm 0.5$}  \\

RAF-DB & 
\textcolor{black}{65.87}\textcolor{black}{\tiny $\pm 1.0$} &  
80.47\textcolor{black}{\tiny $\pm 0.7$} & 
\textcolor{black}{76.85}\textcolor{black}{\tiny $\pm 0.5$} & 
\textcolor{black}{81.75}\textcolor{black}{\tiny $\pm 0.4$}   \\

AffectNet & 
\textcolor{black}{40.28}\textcolor{black}{\tiny $\pm 1.2$} & 
54.91\textcolor{black}{\tiny $\pm 1.0$} &  
\textcolor{black}{51.25}\textcolor{black}{\tiny $\pm 0.6$} & 
\textcolor{black}{55.56}\textcolor{black}{\tiny $\pm 0.5$} \\

\bottomrule
\end{tabular}
\label{tab:iid_summary}
\end{center}
\end{table}

\begin{table*}[ht]
    \centering
    \setlength
    \tabcolsep{3.6pt}
    \caption{The performance of different semi-supervised methods with OOD unlabelled data on FER13 and RAF-DB, when 10, 25, 100, and 250 labelled samples per class are used for training. }
    \begin{tabular}{l|rrrr|rrrr| c}
    \toprule
    & \multicolumn{4}{c|}{\textbf{FER13}} & \multicolumn{4}{c}{\textbf{RAF-DB}} & Avg. Acc.   \\
    \cmidrule(l{3pt}r{3pt}){1-10}
    Method / $m$                & 10 labels        & 25 labels      & 100 labels     & 250 labels    & 10 labels         & 25 labels      & 100 labels     & 250 labels       \\
   \cmidrule(l{3pt}r{3pt}){1-10}
    Pi-Model \cite{pi_model} &
    24.71\textcolor{black}{\tiny $\pm 3.8$} &
    29.02\textcolor{black}{\tiny $\pm 2.6$} &
    43.63\textcolor{black}{\tiny $\pm 1.9$} &
    53.58\textcolor{black}{\tiny $\pm 1.1$} &

    38.62\textcolor{black}{\tiny $\pm 3.1$} &
    39.73\textcolor{black}{\tiny $\pm 2.5$} &
    46.81\textcolor{black}{\tiny $\pm 1.7$} &
    66.53\textcolor{black}{\tiny $\pm 0.6$} &
    42.83\\

    Mean Teacher \cite{mean_teacher} &
    25.16\textcolor{black}{\tiny $\pm 2.8$} &
    27.61\textcolor{black}{\tiny $\pm 1.9$} &
    47.90\textcolor{black}{\tiny $\pm 1.6$} &
    54.82\textcolor{black}{\tiny $\pm 1.0$} &

    38.62\textcolor{black}{\tiny $\pm 2.5$} &
    37.65\textcolor{black}{\tiny $\pm 2.0$} &
    52.09\textcolor{black}{\tiny $\pm 1.5$} &
    65.45\textcolor{black}{\tiny $\pm 0.7$} &
    43.66\\

    VAT \cite{vat} &
    23.95\textcolor{black}{\tiny $\pm 3.1$} &
    28.49\textcolor{black}{\tiny $\pm 2.0$} &
    43.37\textcolor{black}{\tiny $\pm 1.7$} &
    52.98\textcolor{black}{\tiny $\pm 1.1$} &

    38.62\textcolor{black}{\tiny $\pm 2.6$} &
    38.85\textcolor{black}{\tiny $\pm 2.1$} &
    49.58\textcolor{black}{\tiny $\pm 1.6$} &
    61.60\textcolor{black}{\tiny $\pm 0.9$} &
    42.18\\

    Pseudo-label \cite{pseudo_labels} &
    24.09\textcolor{black}{\tiny $\pm 3.0$} &
    36.04\textcolor{black}{\tiny $\pm 1.8$} &
    47.20\textcolor{black}{\tiny $\pm 1.5$} &
    53.23\textcolor{black}{\tiny $\pm 1.0$} &

    38.62\textcolor{black}{\tiny $\pm 2.6$} &
    38.62\textcolor{black}{\tiny $\pm 2.0$} &
    49.84\textcolor{black}{\tiny $\pm 1.6$} &
    63.17\textcolor{black}{\tiny $\pm 0.8$} &
    43.85\\

    UDA \cite{uda} &
    27.60\textcolor{black}{\tiny $\pm 2.7$} &
    38.20\textcolor{black}{\tiny $\pm 1.6$} &
    52.16\textcolor{black}{\tiny $\pm 1.4$} &
    56.13\textcolor{black}{\tiny $\pm 1.0$} &

    39.37\textcolor{black}{\tiny $\pm 2.9$} &
    41.79\textcolor{black}{\tiny $\pm 2.3$} &
    60.95\textcolor{black}{\tiny $\pm 1.3$} &
    65.71\textcolor{black}{\tiny $\pm 0.7$} &
    47.74\\

    MixMatch \cite{mixmatch} &
    31.64\textcolor{black}{\tiny $\pm 2.6$} &
    37.66\textcolor{black}{\tiny $\pm 1.7$} &
    49.68\textcolor{black}{\tiny $\pm 1.2$} &
    56.39\textcolor{black}{\tiny $\pm 0.9$} &

    43.02\textcolor{black}{\tiny $\pm 2.0$} &
    51.14\textcolor{black}{\tiny $\pm 1.6$} &
    63.04\textcolor{black}{\tiny $\pm 1.1$} &
    70.37\textcolor{black}{\tiny $\pm 0.5$} &
    50.37\\

    ReMixMatch \cite{remixmatch} &
    \textbf{33.31}\textcolor{black}{\tiny $\pm 1.9$} &
    \textbf{44.15}\textcolor{black}{\tiny $\pm 1.2$} &
    \underline{53.15}\textcolor{black}{\tiny $\pm 1.0$} &
    \textbf{58.48}\textcolor{black}{\tiny $\pm 0.8$} &

    \textbf{48.37}\textcolor{black}{\tiny $\pm 1.1$} &
    \textbf{58.47}\textcolor{black}{\tiny $\pm 0.9$} &
    67.47\textcolor{black}{\tiny $\pm 0.5$} &
    \textbf{74.87}\textcolor{black}{\tiny $\pm 0.5$} &
    \textbf{54.73}\\

    FixMatch \cite{fixmatch} &
    29.60\textcolor{black}{\tiny $\pm 2.5$} &
    38.98\textcolor{black}{\tiny $\pm 1.6$} &
    52.44\textcolor{black}{\tiny $\pm 1.1$} &
    57.40\textcolor{black}{\tiny $\pm 0.8$} &

    23.99\textcolor{black}{\tiny $\pm 2.8$} &
    45.86\textcolor{black}{\tiny $\pm 1.9$} &
    61.15\textcolor{black}{\tiny $\pm 1.2$} &
    64.80\textcolor{black}{\tiny $\pm 0.6$} &
    46.78\\

    FlexMatch \cite{flexmatch} &
    \underline{31.89}\textcolor{black}{\tiny $\pm 2.1$} &
    \underline{41.0}\textcolor{black}{\tiny $\pm 1.4$} &
    50.43\textcolor{black}{\tiny $\pm 1.2$} &
    56.17\textcolor{black}{\tiny $\pm 0.8$} &

    46.87\textcolor{black}{\tiny $\pm 1.2$} &
    53.16\textcolor{black}{\tiny $\pm 1.0$} &
    \textbf{69.59}\textcolor{black}{\tiny $\pm 0.5$} &
    \underline{72.59}\textcolor{black}{\tiny $\pm 0.3$} &
    52.71\\

    CCSSL \cite{ccssl} &
    30.05\textcolor{black}{\tiny $\pm 2.3$} &
    39.09\textcolor{black}{\tiny $\pm 1.5$} &
    \textbf{53.32}\textcolor{black}{\tiny $\pm 1.1$} &
    \underline{57.77}\textcolor{black}{\tiny $\pm 0.7$} &

    \underline{47.82}\textcolor{black}{\tiny $\pm 1.1$} &
    \underline{56.52}\textcolor{black}{\tiny $\pm 0.9$} &
    \underline{69.33}\textcolor{black}{\tiny $\pm 0.5$} &
    69.07\textcolor{black}{\tiny $\pm 0.3$} &
    \underline{52.87}\\

    \bottomrule
    \end{tabular}
    \label{tab:ood_main}
\end{table*}

\subsubsection{Discussion}
Table \ref{tab:iid_summary} provides a summary of the best results obtained for each dataset using 250 labels per class and compares the performances with fully supervised training with an equal amount of labelled samples. Additionally, the table shows the results obtained with fully supervised learning using all the data, including their labels from the original dataset. For FER13, the semi-supervised method achieves an 8.83\% improvement over fully supervised training with the same amount of labelled data (250 labels per class, for a total of 1750 out of 28K images) and is only 2.16\% lower than fully supervised training with all labelled data (28K images). Similarly, for RAF-DB, the semi-supervised method obtains a 10.98\% improvement over fully supervised training with an equal amount of labelled samples (250 labels per class, for a total of 1750 out of 12k images) and is only 3.62\% lower than fully supervised training with all samples being labelled (12K images). Finally, for AffectNet, the semi-supervised method achieves a 10.98\% improvement over the fully supervised training with the same amount of labelled data (250 labels per class, for a total of, 1750 out of 284k images) and is only 3.66\% lower than the fully supervised training with all labelled data. Based on this summary, we can conclude that semi-supervised methods are able to achieve a significant improvement over fully supervised training with the same amount of labelled data, and can achieve comparable performance to fully supervised training on large amounts of labelled samples. \textcolor{black}{Finally, we show the results for using all the labelled data along with ID unlabelled data. While the unlabelled set is the same size as the labelled set, training in a semi-supervised setting provides improvement over fully-supervised learning.}

\begin{figure}[ht]
    \centering

     \begin{subfigure}[b]{0.15\textwidth}
         \centering
         \includegraphics[width=1.1\textwidth]{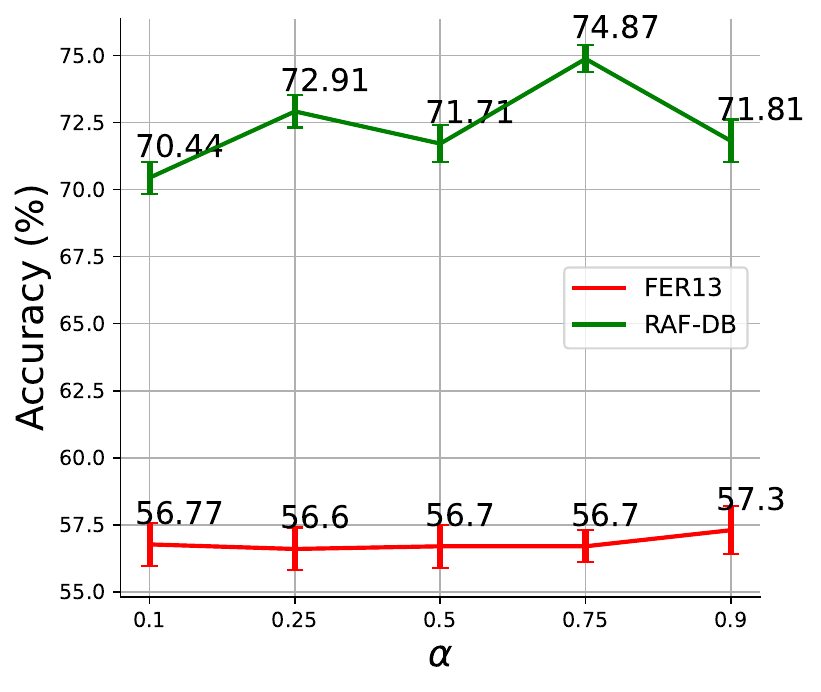}
         \caption{Accuracy vs. ${\alpha}$ for \textit{\textbf{ReMixMatch}}.}
         \label{fig:ood_alpha_remixmatch}
     \end{subfigure}
      \hspace{5pt}
      \begin{subfigure}[b]{0.15\textwidth}
         \centering
         \includegraphics[width=1.1\textwidth]{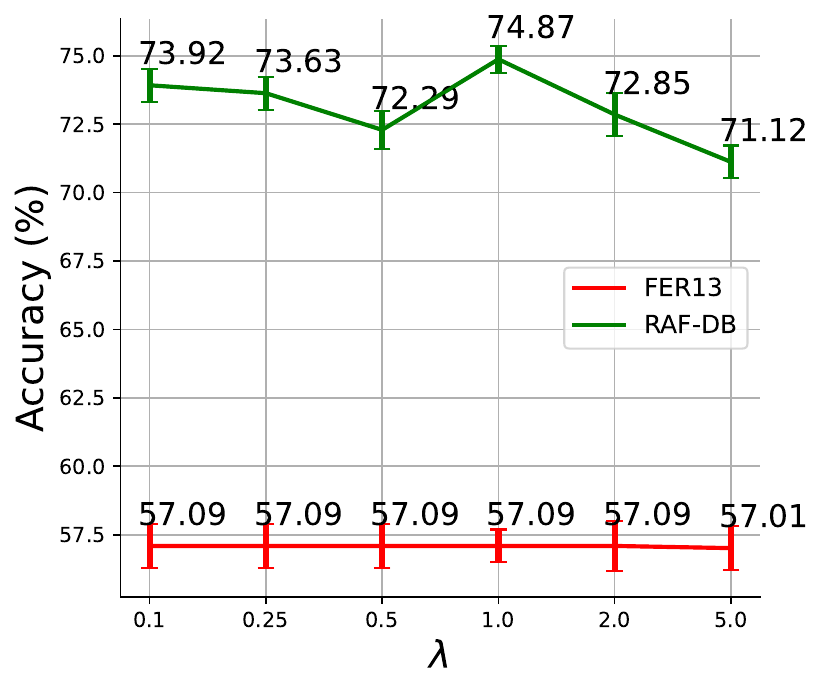}
         \caption{Accuracy vs. $\lambda$ for \textit{\textbf{ReMixMatch}}.}
        \label{fig:ood_lambda_remixmatch}
     \end{subfigure}
     \\
      \begin{subfigure}[b]{0.15\textwidth}
         \centering
         \includegraphics[width=1.1\textwidth]{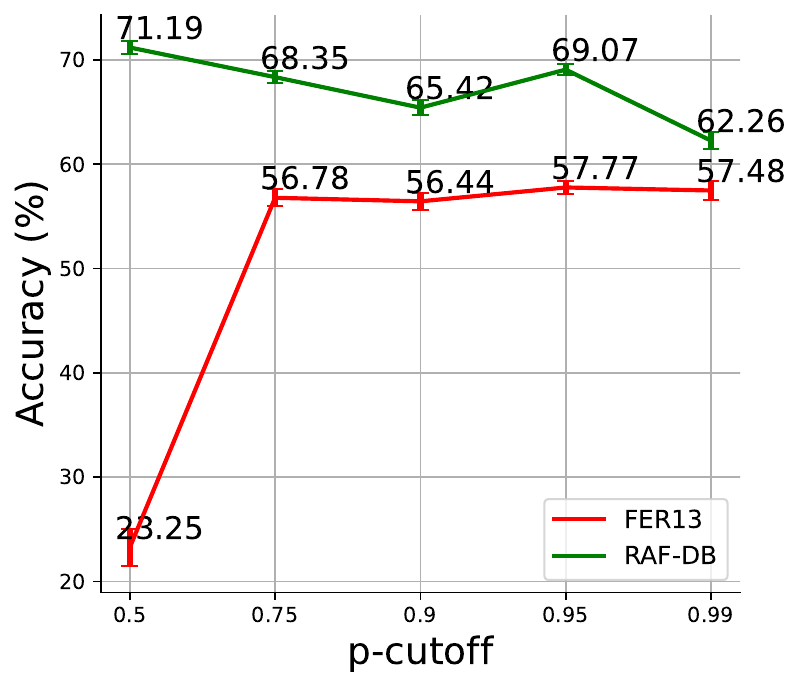}
         \caption{Accuracy vs. $P_\text{cutoff}$ for \textit{\textbf{CCSSL}}.}
         \label{fig:ood_cutoff_ccssl}
     \end{subfigure}
     \hspace{5pt}
      \begin{subfigure}[b]{0.15\textwidth}
         \centering
         \includegraphics[width=1.1\textwidth]{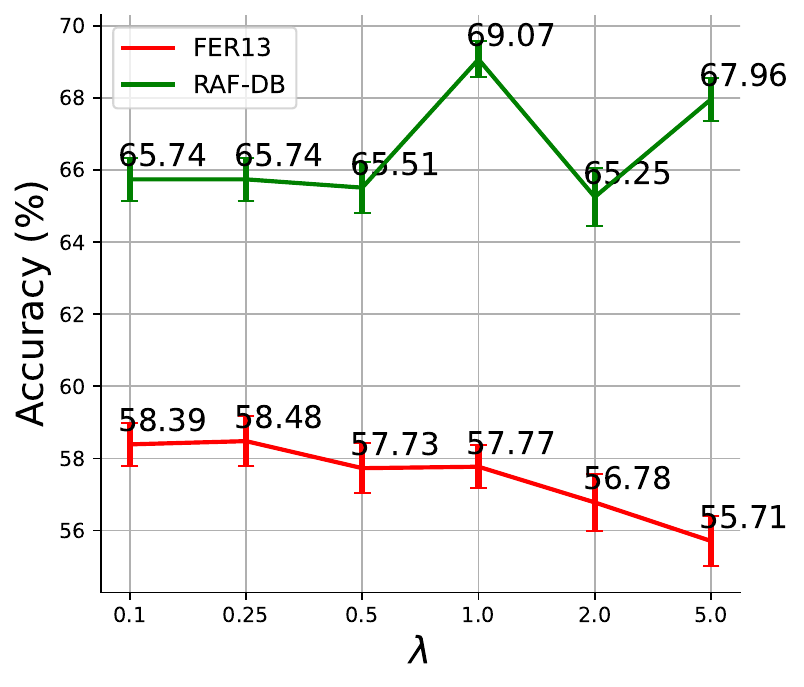}
         \caption{Accuracy vs. $\lambda$ for \textit{\textbf{CCSSL}}.}
         \label{fig:ood_lambda_ccssl}
     \end{subfigure}

    \caption{Sensitivity study of various parameters for two of the best semi-supervised methods on OOD unlabelled data. }
    \label{fig:sensitivity_ood}
\end{figure}

\begin{table}[]
\setlength
\tabcolsep{4.0pt}
\caption{Comparison between fully-supervised learning and semi-supervised learning with OOD unlabelled data. }
\begin{center}
\begin{tabular}{l|cc | ccc}
\toprule
    & \multicolumn{2}{c|}{\textbf{Supervised}} & \multicolumn{3}{c}{\textbf{Semi-sup.}} \\
    
    \cmidrule(l{3pt}r{3pt}){1-6}
    
    \textbf{Dataset} & \textbf{All data}  & \textbf{ID/250} & \textbf{ID/250} & \textbf{OOD/250} & \textcolor{black}{\textbf{OOD/All}} \\
    
    \cmidrule(l{3pt}r{3pt}){1-6}
    
    FER13 & 
    64.57\textcolor{black}{\tiny $\pm 0.9$} & 
    \textcolor{black}{53.58}\textcolor{black}{\tiny $\pm 1.1$} &  
    \textcolor{black}{62.20}\textcolor{black}{\tiny $\pm 0.5$} &  
    \textcolor{black}{58.48}\textcolor{black}{\tiny $\pm 0.8$} & 
    \textcolor{black}{70.40}\textcolor{black}{\tiny $\pm 0.4$}\\
    
    RAF-DB & 
    80.47\textcolor{black}{\tiny $\pm 0.7$} & 
    \textcolor{black}{65.87}\textcolor{black}{\tiny $\pm 1.0$} & 
    \textcolor{black}{76.85}\textcolor{black}{\tiny $\pm 0.5$}&
    \textcolor{black}{74.87}\textcolor{black}{\tiny $\pm 0.5$} & 
    \textcolor{black}{83.73}\textcolor{black}{\tiny $\pm 0.2$} \\

\bottomrule
\end{tabular}
\label{tab:ood_summary}
\end{center}
\end{table}

\begin{table*}[ht]
    \centering
    \setlength
    \tabcolsep{1.6pt}
    \caption{The performance of different semi-supervised methods with unconstrained unlabelled data on FER13, RAF-DB, and AffectNet, when 10, 25, 100, and 250 labelled samples per class are used for training. }
    \resizebox{1\textwidth}{!}
    {
    \begin{tabular}{l|rrrr|rrrr|rrrr| c}
    \toprule
    & \multicolumn{4}{c|}{\textbf{FER13}} & \multicolumn{4}{c|}{\textbf{RAF-DB}}  & \multicolumn{4}{c|}{\textbf{AffectNet}}  & Avg. Acc\\
    \cmidrule(l{3pt}r{3pt}){1-14}
    Method / $m$                & 10 labels        & 25 labels      & 100 labels     & 250 labels    & 10 labels         & 25 labels      & 100 labels     & 250 labels    & 10 labels         & 25 labels      & 100 labels     & 250 labels      \\
   \cmidrule(l{3pt}r{3pt}){1-14}
    Pi-Model \cite{pi_model} &
    22.28\textcolor{black}{\tiny $\pm 3.8$} &
    27.03\textcolor{black}{\tiny $\pm 2.5$} &
    44.96\textcolor{black}{\tiny $\pm 1.5$} &
    54.79\textcolor{black}{\tiny $\pm 1.0$} &

    38.62\textcolor{black}{\tiny $\pm 3.1$} &
    38.62\textcolor{black}{\tiny $\pm 3.0$} &
    52.18\textcolor{black}{\tiny $\pm 1.5$} &
    65.74\textcolor{black}{\tiny $\pm 0.6$} &

    18.00\textcolor{black}{\tiny $\pm 4.3$} &
    18.74\textcolor{black}{\tiny $\pm 3.2$} &
    19.66\textcolor{black}{\tiny $\pm 2.8$} &
    35.57\textcolor{black}{\tiny $\pm 1.7$} &
    36.35\\

    Mean Teacher \cite{mean_teacher} &
    24.88\textcolor{black}{\tiny $\pm 3.0$} &
    29.77\textcolor{black}{\tiny $\pm 2.0$} &
    43.66\textcolor{black}{\tiny $\pm 1.6$} &
    53.22\textcolor{black}{\tiny $\pm 1.1$} &

    36.18\textcolor{black}{\tiny $\pm 3.1$} &
    38.62\textcolor{black}{\tiny $\pm 2.9$} &
    50.59\textcolor{black}{\tiny $\pm 1.6$} &
    66.49\textcolor{black}{\tiny $\pm 0.6$} &

    17.69\textcolor{black}{\tiny $\pm 4.3$} &
    19.11\textcolor{black}{\tiny $\pm 3.1$} &
    21.34\textcolor{black}{\tiny $\pm 2.7$} &
    38.34\textcolor{black}{\tiny $\pm 1.4$} &
    36.66\\

    VAT \cite{vat} &
    21.94\textcolor{black}{\tiny $\pm 3.0$} &
    30.05\textcolor{black}{\tiny $\pm 2.0$} &
    43.56\textcolor{black}{\tiny $\pm 1.6$} &
    52.42\textcolor{black}{\tiny $\pm 1.1$} &

    38.62\textcolor{black}{\tiny $\pm 3.1$} &
    41.17\textcolor{black}{\tiny $\pm 2.5$} &
    51.76\textcolor{black}{\tiny $\pm 1.6$} &
    62.32\textcolor{black}{\tiny $\pm 0.8$} &

    17.86\textcolor{black}{\tiny $\pm 3.3$} &
    19.23\textcolor{black}{\tiny $\pm 3.1$} &
    20.89\textcolor{black}{\tiny $\pm 2.7$} &
    26.34\textcolor{black}{\tiny $\pm 1.8$} &
    35.51\\

    Pseudo-label \cite{pseudo_labels} &
    23.36\textcolor{black}{\tiny $\pm 3.0$} &
    33.51\textcolor{black}{\tiny $\pm 1.8$} &
    47.70\textcolor{black}{\tiny $\pm 1.5$} &
    53.57\textcolor{black}{\tiny $\pm 1.0$} &

    38.62\textcolor{black}{\tiny $\pm 3.1$} &
    37.45\textcolor{black}{\tiny $\pm 2.0$} &
    49.54\textcolor{black}{\tiny $\pm 1.6$} &
    63.40\textcolor{black}{\tiny $\pm 0.8$} &

    17.29\textcolor{black}{\tiny $\pm 3.3$} &
    19.54\textcolor{black}{\tiny $\pm 3.0$} &
    21.06\textcolor{black}{\tiny $\pm 2.6$} &
    24.03\textcolor{black}{\tiny $\pm 1.7$} &
    35.76\\

    UDA \cite{uda} &
    24.88\textcolor{black}{\tiny $\pm 2.9$} &
    33.99\textcolor{black}{\tiny $\pm 1.8$} &
    49.40\textcolor{black}{\tiny $\pm 1.4$} &
    52.97\textcolor{black}{\tiny $\pm 1.0$} &

    27.05\textcolor{black}{\tiny $\pm 3.0$} &
    43.94\textcolor{black}{\tiny $\pm 1.9$} &
    52.41\textcolor{black}{\tiny $\pm 1.5$} &
    62.87\textcolor{black}{\tiny $\pm 0.7$} &

    16.91\textcolor{black}{\tiny $\pm 3.9$} &
    17.60\textcolor{black}{\tiny $\pm 3.0$} &
    28.31\textcolor{black}{\tiny $\pm 1.9$} &
    33.14\textcolor{black}{\tiny $\pm 1.6$} &
    36.96\\

    MixMatch \cite{mixmatch} &
    31.43\textcolor{black}{\tiny $\pm 2.7$} &
    39.44\textcolor{black}{\tiny $\pm 1.7$} &
    \underline{51.99}\textcolor{black}{\tiny $\pm 1.2$} &
    \underline{56.26}\textcolor{black}{\tiny $\pm 0.9$} &

    \underline{42.37}\textcolor{black}{\tiny $\pm 1.1$} &
    \underline{50.52}\textcolor{black}{\tiny $\pm 0.9$} &
    60.01\textcolor{black}{\tiny $\pm 0.5$} &
    \underline{68.84}\textcolor{black}{\tiny $\pm 0.4$} &

    \underline{25.14}\textcolor{black}{\tiny $\pm 3.0$} &
    26.26\textcolor{black}{\tiny $\pm 2.0$} &
    \underline{33.40}\textcolor{black}{\tiny $\pm 1.6$} &
    \underline{39.23}\textcolor{black}{\tiny $\pm 1.2$} &
    \underline{43.74}\\

    ReMixMatch \cite{remixmatch} &
    \underline{32.49}\textcolor{black}{\tiny $\pm 2.3$} &
    \textbf{42.05}\textcolor{black}{\tiny $\pm 1.4$} &
    \textbf{52.83}\textcolor{black}{\tiny $\pm 1.1$} &
    \textbf{58.04}\textcolor{black}{\tiny $\pm 0.8$} &

    \textbf{49.15}\textcolor{black}{\tiny $\pm 1.0$} &
    \textbf{54.60}\textcolor{black}{\tiny $\pm 0.8$} &
    \textbf{64.50}\textcolor{black}{\tiny $\pm 0.5$} &
    \textbf{70.70}\textcolor{black}{\tiny $\pm 0.5$} &

    \textbf{26.29}\textcolor{black}{\tiny $\pm 2.7$} &
    \textbf{29.23}\textcolor{black}{\tiny $\pm 1.8$} &
    \textbf{37.31}\textcolor{black}{\tiny $\pm 1.4$} &
    \textbf{40.40}\textcolor{black}{\tiny $\pm 1.0$} &
    \textbf{46.47}\\

    FixMatch \cite{fixmatch} &
    26.58\textcolor{black}{\tiny $\pm 2.6$} &
    35.16\textcolor{black}{\tiny $\pm 1.7$} &
    49.29\textcolor{black}{\tiny $\pm 1.3$} &
    54.30\textcolor{black}{\tiny $\pm 1.0$} &

    33.41\textcolor{black}{\tiny $\pm 2.9$} &
    39.99\textcolor{black}{\tiny $\pm 1.9$} &
    52.93\textcolor{black}{\tiny $\pm 1.4$} &
    60.59\textcolor{black}{\tiny $\pm 0.7$} &

    15.77\textcolor{black}{\tiny $\pm 3.9$} &
    17.23\textcolor{black}{\tiny $\pm 3.0$} &
    28.46\textcolor{black}{\tiny $\pm 1.9$} &
    31.23\textcolor{black}{\tiny $\pm 1.6$} &
    37.08\\

    FlexMatch \cite{flexmatch} &
    25.42\textcolor{black}{\tiny $\pm 2.7$} &
    37.21\textcolor{black}{\tiny $\pm 1.7$} &
    50.21\textcolor{black}{\tiny $\pm 1.3$} &
    54.83\textcolor{black}{\tiny $\pm 1.0$} &

    36.86\textcolor{black}{\tiny $\pm 2.8$} &
    49.45\textcolor{black}{\tiny $\pm 1.8$} &
    59.62\textcolor{black}{\tiny $\pm 1.2$} &
    64.05\textcolor{black}{\tiny $\pm 0.7$} &

    24.40\textcolor{black}{\tiny $\pm 3.3$} &
    \underline{28.06}\textcolor{black}{\tiny $\pm 1.8$} &
    32.26\textcolor{black}{\tiny $\pm 1.7$} &
    36.74\textcolor{black}{\tiny $\pm 1.2$} &
    41.59\\

    CoMatch \cite{comatch} &
    \textbf{33.01}\textcolor{black}{\tiny $\pm 2.6$} &
    \underline{40.90}\textcolor{black}{\tiny $\pm 1.6$} &
    50.67\textcolor{black}{\tiny $\pm 1.3$} &
    55.83\textcolor{black}{\tiny $\pm 1.0$} &

    27.95\textcolor{black}{\tiny $\pm 3.0$} &
    32.53\textcolor{black}{\tiny $\pm 2.0$} &
    \underline{64.11}\textcolor{black}{\tiny $\pm 1.1$} &
    68.17\textcolor{black}{\tiny $\pm 0.6$} &

    23.40\textcolor{black}{\tiny $\pm 3.4$} &
    27.50\textcolor{black}{\tiny $\pm 2.0$} &
    32.33\textcolor{black}{\tiny $\pm 1.8$} &
    35.79\textcolor{black}{\tiny $\pm 1.3$} &
    41.02\\

    CCSSL \cite{ccssl} &
    24.69\textcolor{black}{\tiny $\pm 2.8$} &
    39.51\textcolor{black}{\tiny $\pm 1.7$} &
    51.14\textcolor{black}{\tiny $\pm 1.3$} &
    55.63\textcolor{black}{\tiny $\pm 1.0$} &

    31.32\textcolor{black}{\tiny $\pm 2.9$} &
    47.26\textcolor{black}{\tiny $\pm 1.8$} &
    55.25\textcolor{black}{\tiny $\pm 1.3$} &
    61.90\textcolor{black}{\tiny $\pm 0.7$} &

    16.51\textcolor{black}{\tiny $\pm 2.3$} &
    25.11\textcolor{black}{\tiny $\pm 1.9$} &
    25.00\textcolor{black}{\tiny $\pm 1.7$} &
    33.71\textcolor{black}{\tiny $\pm 1.3$} &
    38.92\\

    \bottomrule
    \end{tabular}
    }
    \label{tab:unconstrained_main}
\end{table*}

\subsection{Semi-supervised FER with OOD unlabelled data}\label{subsec:semi-supervised-fer-with-OOD-unlabelled-data}

\subsubsection{Setup} 
\textcolor{black}{
In this section, we present the results of OOD semi-supervised learning. In this setting, the unlabelled data consists of images belonging to the same expression categories as the labelled data but originating from different sources and therefore having a different distribution. Specifically, we use a pre-defined number of samples (10, 25, 100, or 250 labelled samples per class) from the training set of FER-13 and RAF-DB as the labelled data and the complete training set (images without labels) of AffectNet as unlabelled (OOD) data. The accuracy is reported on the validation set of the FER-13 and RAF-DB datasets, respectively. 
}

\subsubsection{Performance}
The performance of different semi-supervised methods with OOD unlabelled data are presented in Table \ref{tab:ood_main}. We can draw two key observations from this table. Firstly, we can observe a significant drop in the performance of all methods compared to when ID unlabelled data are used. These findings are consistent with previous works in the OOD semi-supervised literature \cite{oliver2018realistic, su2021realistic}. For instance, in our study, we find that FixMatch achieves a 62.20\% accuracy on FER-13 (250 labels) with ID unlabelled data, but this drops to 57.4\% when OOD data are utilized. The performance drop is even more substantial with smaller labelled set sizes. For example, when only 10 samples per class are available on FER-13, the FixMatch performance drops from 47.88\% to only 29.6\%. The second observation is that the best-performing method is different for OOD semi-supervised learning. While FixMatch and MixMatch were the top two methods for ID semi-supervised learning, ReMixMatch, and CCSSL are the top two methods for OOD semi-supervised learning. Both of these methods perform significantly better than FixMatch.

\subsubsection{Sensitivity study} 
Next, we conduct a sensitivity analysis on the two best-performing methods, ReMixMatch and CCSSL, for OOD semi-supervised learning. Since these methods were originally designed for ID semi-supervised learning in general computer vision applications, it is crucial to explore their performance for different hyperparameters under the OOD setting.
Figure \ref{fig:sensitivity_ood} presents the results of this study. We examine the impact of $\alpha$ and $\lambda$ for ReMixMatch and $P_\text{cutoff}$ and $\lambda$ for CCSSL. In the experiment with ReMixMatch's $\alpha$ values (Figure \ref{fig:ood_alpha_remixmatch}), we discover that the best results for both datasets are achieved with a value of $0.75$, which is also the default in the original ReMixMatch paper. We also find that the optimal value for $\lambda$ was $1.0$ for both datasets. In the experiments on CCSSL's $P_\text{cutoff}$ value (Figure \ref{fig:ood_cutoff_ccssl}), we observe that lower values of this parameter yields better results. Specifically, RAF-DB produces the best result with a value of \textit{0.5}, while FER13 achieves the best result with a value of \textit{0.75}. Finally, for $\lambda$ values (Fig \ref{fig:ood_lambda_ccssl}), we obtain the best accuracy for different values for RAF-DB and FER13 datasets, with \textit{1.0} and \textit{0.25} respectively.

\subsubsection{Discussion}
Table \ref{tab:ood_summary} summarizes the results of semi-supervised FER using OOD unlabelled data and compares them to both fully-supervised and semi-supervised methods with ID unlabelled data. All results are shown for 250 labelled samples per class. We observe a drop in performance of 1.98\% and 3.72\% for RAF-DB and FER13 datasets, respectively, compared to ID semi-supervised learning. However, the performance of semi-supervised learning with OOD unlabelled samples is still better than fully-supervised learning by 9.0\% and 4.9\% for the two datasets, respectively. Therefore, we can conclude that learning with the presence of OOD data is still a better choice than relying on a fully supervised setting alone when presented with limited labelled data. \textcolor{black}{Finally, we find that using all the labelled data along with the OOD unlabelled data provides considerable improvement (5.83\% and 3.26\% for FER-13 and RAF-DB) over fully supervised learning with all the labelled data. }

\begin{figure}[ht]
    \centering

     \begin{subfigure}[b]{0.15\textwidth}
         \centering
         \includegraphics[width=1.1\textwidth]{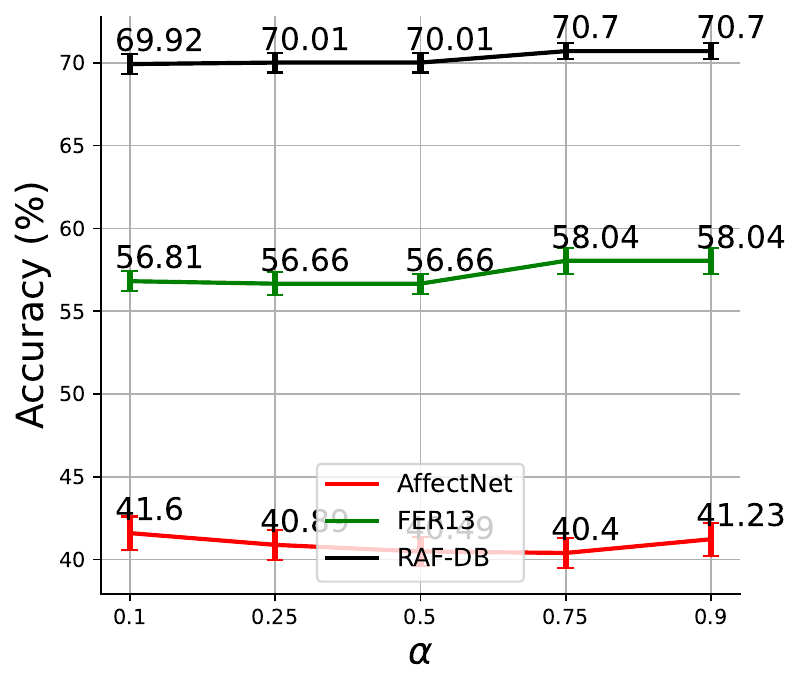}
         \caption{Accuracy vs. ${\alpha}$ for \textit{\textbf{ReMixMatch}}.}
         \label{fig:unc_alpha_remixmatch}
     \end{subfigure}
     \hspace{5pt}
     \begin{subfigure}[b]{0.15\textwidth}
         \centering
         \includegraphics[width=1.1\textwidth]{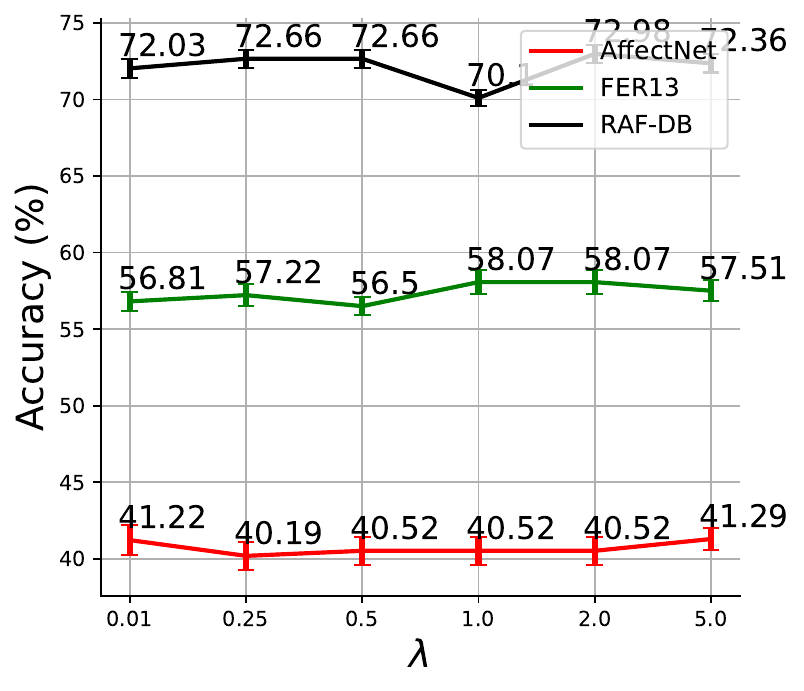}
         \caption{Accuracy vs. $\lambda$ for \textit{\textbf{ReMixMatch}}.}
        \label{fig:unc_lambda_remixmatch}
     \end{subfigure}
     \\
     \begin{subfigure}[b]{0.15\textwidth}
         \centering
         \includegraphics[width=1.1\textwidth]{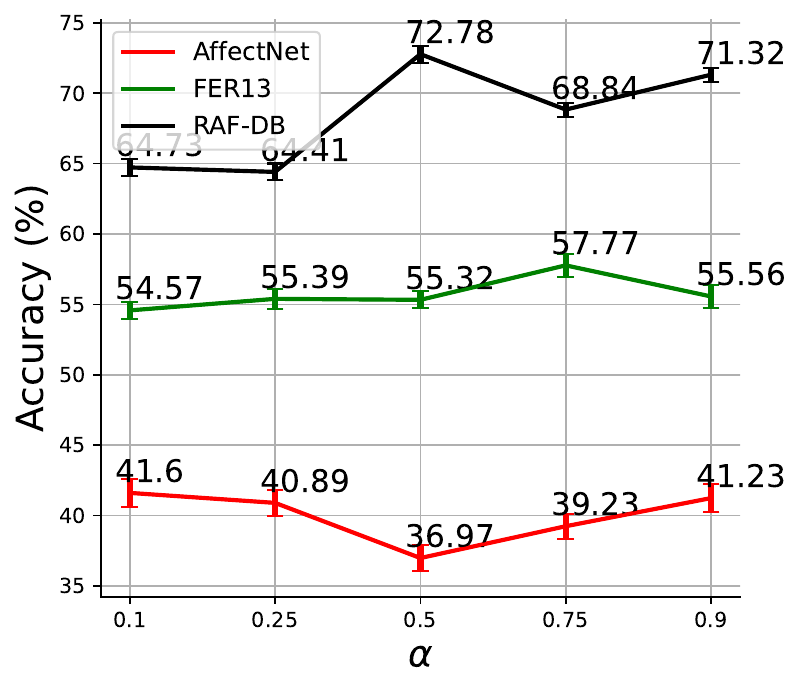}
         \caption{Accuracy vs. $\alpha$ for \textit{\textbf{MixMatch}}.}
         \label{fig:unc_alpha_mixmatch}
     \end{subfigure}
     \hspace{5pt}
      \begin{subfigure}[b]{0.15\textwidth}
         \centering
         \includegraphics[width=1.1\textwidth]{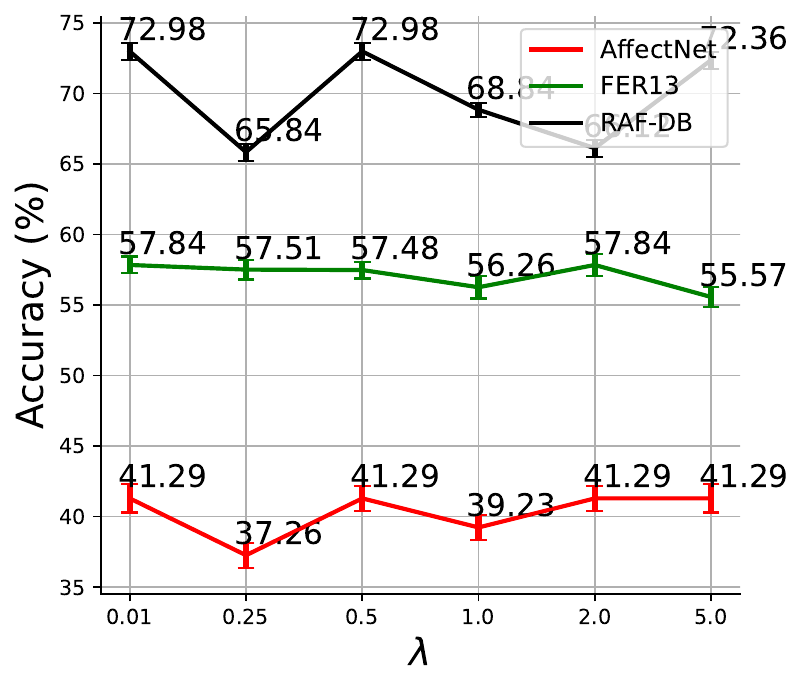}
         \caption{Accuracy vs. $\lambda$ for \textit{\textbf{MixMatch}}.}
         \label{fig:unc_lambda_mixmatch}
     \end{subfigure}

    \caption{Sensitivity study of various parameters for two of the best semi-supervised methods on unconstrained unlabelled data. }
    \label{fig:sensitivity_unconstrained}
\end{figure}

\begin{table}[]
\setlength
\tabcolsep{4.0pt}
\caption{Comparison between fully-supervised learning and semi-supervised learning with unconstrained unlabelled data. }
\begin{center}
\begin{tabular}{l|cc | cc c}
\toprule

& \multicolumn{2}{c|}{\textbf{Supervised}} & \multicolumn{3}{c}{\textbf{Semi-sup.}} \\
\cmidrule(l{3pt}r{3pt}){1-6}

\textbf{Dataset} & \textbf{All data}  & \textbf{ID/250} & \textbf{ID/250} & \textbf{Unc./250} & \textcolor{black}{\textbf{Unc./All}} \\

    \cmidrule(l{3pt}r{3pt}){1-6}

    FER13 & 
    64.57\textcolor{black}{\tiny $\pm 0.9$} & 
    \textcolor{black}{53.58}\textcolor{black}{\tiny $\pm 1.1$} &  
    \textcolor{black}{62.20}\textcolor{black}{\tiny $\pm 0.5$} &  
    \textcolor{black}{58.04}\textcolor{black}{\tiny $\pm 0.8$} & 
    \textcolor{black}{70.50}\textcolor{black}{\tiny $\pm 0.4$}\\
    
    RAF-DB & 
    80.47\textcolor{black}{\tiny $\pm 0.7$} & 
    \textcolor{black}{65.87}\textcolor{black}{\tiny $\pm 1.0$} &
    \textcolor{black}{76.85}\textcolor{black}{\tiny $\pm 0.5$}& \textcolor{black}{70.70}\textcolor{black}{\tiny $\pm 0.5$} & 
    \textcolor{black}{82.75}\textcolor{black}{\tiny $\pm 0.2$} \\
    
    AffectNet & 
    54.91\textcolor{black}{\tiny $\pm 1.0$} & 
    \textcolor{black}{40.28}\textcolor{black}{\tiny $\pm 1.2$} & 
    \textcolor{black}{51.25}\textcolor{black}{\tiny $\pm 0.6$} & 
    \textcolor{black}{40.40}\textcolor{black}{\tiny $\pm 1.0$} & 
    \textcolor{black}{57.65}\textcolor{black}{\tiny $\pm 0.6$}\\
    \bottomrule
\end{tabular}
\label{tab:unconstrained_summary}
\end{center}
\end{table}

\subsection{Semi-supervised FER with \textit{unconstrained} unlabelled data}

\subsubsection{Setup}
\textcolor{black}{
In this section, we present the results of semi-supervised FER using unconstrained unlabelled data, which is considered the most challenging setting for semi-supervised learning. Here, the unlabelled data are obtained from a different source than the labelled data and do not necessarily contain images of known classes. In unconstrained semi-supervised learning, we use a pre-defined number of samples (10, 25, 100, or 250 labelled samples per class) from the training set of FER-13, RAF-DB, and AffectNet as the labelled data and complete training set of AffectNet (do not contain labels) as unlabelled data. The accuracy is reported on the validation set of the FER-13, RAF-DB, and AffectNet, respectively.
}

\subsubsection{Performance}
Table \ref{tab:unconstrained_main} shows the results of different semi-supervised methods with unconstrained unlabelled data. We make the following two observations from this table. Firstly, The best-performing method on unconstrained unlabelled data is similar to that of the OOD setting. Again, ReMixMatch achieves the best average results compared to the other methods. However, the second-best method is different from the OOD setting (CCSSL), and is now MixMatch. The performance of the rest of the methods is significantly lower than these two methods. Secondly, the performance of semi-supervised methods is much lower than training with ID data but not too far away from OOD data. For example, the performance of ReMixMatch on FER-13 with just 10 labelled samples on unconstrained unlabelled data is 8.58\% lower than training with ID labelled data, but only 0.82\% lower than on OOD data. This indicates that \textbf{for FER, unconstrained unlabelled data can result in competitive semi-supervised performance to OOD samples.}
This is an important finding since it is more convenient, practical, and economical to collect large unconstrained unlabelled data than ID data or even OOD data of expressive faces.

\begin{table*}[t]
    \centering
    \setlength
    \tabcolsep{3.6pt}
    \caption{The performance of different semi-supervised methods with small unlabelled data on KDEF and DDCF, when 10, 25, 100, and 250 labelled samples per class are used for training. }
    \begin{tabular}{l|rrrr|rrrr|c}
    \toprule
    & \multicolumn{4}{c|}{\textbf{KDEF}} & \multicolumn{4}{c}{\textbf{DDCF}}  & Avg. Acc   \\
    \cmidrule(l{3pt}r{3pt}){1-10}
    Method / $m$                & 10 labels        & 25 labels      & 100 labels     & 250 labels    & 10 labels         & 25 labels      & 100 labels     & 250 labels       \\
    \cmidrule(l{3pt}r{3pt}){1-10}

    Pi-Model \cite{pi_model} & 
    31.90\textcolor{black}{\tiny $\pm 2.0$} &
    58.90\textcolor{black}{\tiny $\pm 1.1$} &
    85.28\textcolor{black}{\tiny $\pm 0.7$} &
    94.27\textcolor{black}{\tiny $\pm 0.3$} &
    
    19.81\textcolor{black}{\tiny $\pm 2.6$} &
    30.82\textcolor{black}{\tiny $\pm 1.8$} &
    77.20\textcolor{black}{\tiny $\pm 0.9$} &
    88.99\textcolor{black}{\tiny $\pm 0.4$} &
    60.90\\
    
    Mean Teacher \cite{mean_teacher} & 
    29.24\textcolor{black}{\tiny $\pm 2.1$} &
    50.31\textcolor{black}{\tiny $\pm 1.3$} &
    82.82\textcolor{black}{\tiny $\pm 0.8$} &
    91.62\textcolor{black}{\tiny $\pm 0.4$} &
    
    18.40\textcolor{black}{\tiny $\pm 2.7$} &
    30.03\textcolor{black}{\tiny $\pm 1.9$} &
    74.37\textcolor{black}{\tiny $\pm 1.0$} &
    89.94\textcolor{black}{\tiny $\pm 0.4$} &
    58.34\\
    
    VAT \cite{vat} & 
    24.54\textcolor{black}{\tiny $\pm 2.2$} &
    44.79\textcolor{black}{\tiny $\pm 1.4$} &
    79.75\textcolor{black}{\tiny $\pm 0.9$} &
    91.21\textcolor{black}{\tiny $\pm 0.7$} &
    
    23.58\textcolor{black}{\tiny $\pm 2.5$} &
    40.57\textcolor{black}{\tiny $\pm 1.8$} &
    72.01\textcolor{black}{\tiny $\pm 1.1$} &
    88.99\textcolor{black}{\tiny $\pm 0.5$} &
    58.18\\
    
    Pseudo-label \cite{pseudo_labels} & 
    31.08\textcolor{black}{\tiny $\pm 2.1$} &
    53.37\textcolor{black}{\tiny $\pm 1.3$} &
    80.98\textcolor{black}{\tiny $\pm 0.8$} &
    93.25\textcolor{black}{\tiny $\pm 0.4$} &
    
    29.40\textcolor{black}{\tiny $\pm 2.5$} &
    41.19\textcolor{black}{\tiny $\pm 1.8$} &
    77.83\textcolor{black}{\tiny $\pm 1.0$} &
    86.79\textcolor{black}{\tiny $\pm 0.7$} &
    61.74\\
    
    UDA \cite{uda} & 
    \textbf{36.81}\textcolor{black}{\tiny $\pm 1.9$} &
    46.42\textcolor{black}{\tiny $\pm 1.2$} &
    88.14\textcolor{black}{\tiny $\pm 0.6$} &
    \underline{97.55}\textcolor{black}{\tiny $\pm 0.2$} &
    
    30.35\textcolor{black}{\tiny $\pm 2.5$} &
    80.82\textcolor{black}{\tiny $\pm 0.9$} &
    86.79\textcolor{black}{\tiny $\pm 0.7$} &
    93.71\textcolor{black}{\tiny $\pm 0.4$} &
    70.07\\
    
    MixMatch \cite{mixmatch} & 
    28.83\textcolor{black}{\tiny $\pm 2.0$} &
    \underline{63.39}\textcolor{black}{\tiny $\pm 1.1$} &
    91.21\textcolor{black}{\tiny $\pm 0.6$} &
    93.87\textcolor{black}{\tiny $\pm 0.4$} &
    
    14.78\textcolor{black}{\tiny $\pm 2.6$} &
    \underline{84.28}\textcolor{black}{\tiny $\pm 0.7$} &
    \underline{89.15}\textcolor{black}{\tiny $\pm 0.5$} &
    93.40\textcolor{black}{\tiny $\pm 0.3$} &
    69.86\\
    
    ReMixMatch \cite{remixmatch} & 
    29.86\textcolor{black}{\tiny $\pm 1.9$} &
    \textbf{67.08}\textcolor{black}{\tiny $\pm 1.0$} &
    \textbf{93.46}\textcolor{black}{\tiny $\pm 0.4$} &
    \textbf{97.75}\textcolor{black}{\tiny $\pm 0.2$} &
    
    25.79\textcolor{black}{\tiny $\pm 2.5$} &
    \textbf{85.53}\textcolor{black}{\tiny $\pm 0.7$} &
    \textbf{89.62}\textcolor{black}{\tiny $\pm 0.5$} &
    \textbf{95.60}\textcolor{black}{\tiny $\pm 0.3$} &
    \textbf{73.09}\\
     
    FixMatch \cite{fixmatch} & 
    \underline{33.54}\textcolor{black}{\tiny $\pm 2.0$} &
    54.60\textcolor{black}{\tiny $\pm 1.2$} &
    89.16\textcolor{black}{\tiny $\pm 0.6$} &
    97.55\textcolor{black}{\tiny $\pm 0.3$} &
    
    38.52\textcolor{black}{\tiny $\pm 2.6$} &
    79.40\textcolor{black}{\tiny $\pm 0.9$} &
    86.64\textcolor{black}{\tiny $\pm 0.8$} &
    \underline{94.65}\textcolor{black}{\tiny $\pm 0.3$} &
    \textbf{71.76}\\
    
    FlexMatch \cite{flexmatch} & 
    27.61\textcolor{black}{\tiny $\pm 2.1$} &
    40.08\textcolor{black}{\tiny $\pm 1.3$} &
    \underline{93.05}\textcolor{black}{\tiny $\pm 0.5$} &
    96.32\textcolor{black}{\tiny $\pm 0.3$} &
    
    \underline{39.62}\textcolor{black}{\tiny $\pm 2.6$} &
    24.37\textcolor{black}{\tiny $\pm 1.9$} &
    86.01\textcolor{black}{\tiny $\pm 0.7$} &
    93.71\textcolor{black}{\tiny $\pm 0.6$} &
    62.60\\
    
    CoMatch \cite{comatch} & 
    21.08\textcolor{black}{\tiny $\pm 2.2$} &
    52.12\textcolor{black}{\tiny $\pm 1.3$} &
    85.05\textcolor{black}{\tiny $\pm 0.7$} &
    92.16\textcolor{black}{\tiny $\pm 0.4$} &
    
    \textbf{44.94}\textcolor{black}{\tiny $\pm 2.6$} &
    63.52\textcolor{black}{\tiny $\pm 1.8$} &
    87.18\textcolor{black}{\tiny $\pm 0.7$} &
    90.70\textcolor{black}{\tiny $\pm 0.5$} &
    67.09\\
    
    CCSSL \cite{ccssl} & 
    19.43\textcolor{black}{\tiny $\pm 2.2$} &
    37.83\textcolor{black}{\tiny $\pm 1.4$} &
    86.50\textcolor{black}{\tiny $\pm 0.9$} &
    94.27\textcolor{black}{\tiny $\pm 0.5$} &
    
    14.15\textcolor{black}{\tiny $\pm 2.7$} &
    55.50\textcolor{black}{\tiny $\pm 1.9$} &
    86.01\textcolor{black}{\tiny $\pm 0.8$} &
    94.65\textcolor{black}{\tiny $\pm 0.4$} &
    61.04\\

    \bottomrule
    \end{tabular}
    \label{tab:small_dataset}
\end{table*}

\begin{figure}[ht]
    \centering

     \begin{subfigure}[b]{0.15\textwidth}
         \centering
         \includegraphics[width=1.1\textwidth]{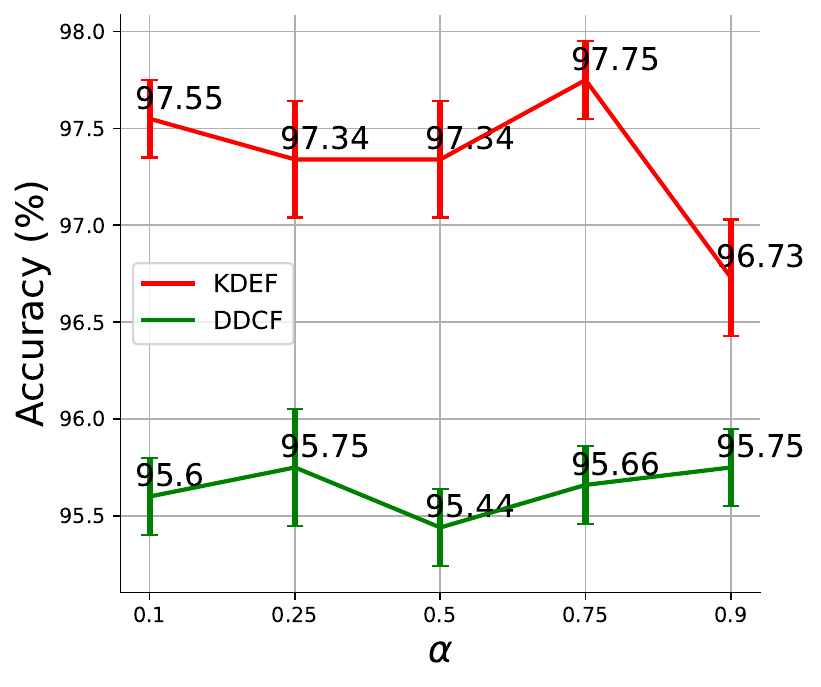}
         \caption{Acc. vs. ${\alpha}$ for \textit{\textbf{ReMixMatch}}.}
         \label{fig:small_alpha_remixmatch}
     \end{subfigure}
     \hspace{5pt}
     \begin{subfigure}[b]{0.15\textwidth}
         \centering
         \includegraphics[width=1.1\textwidth]{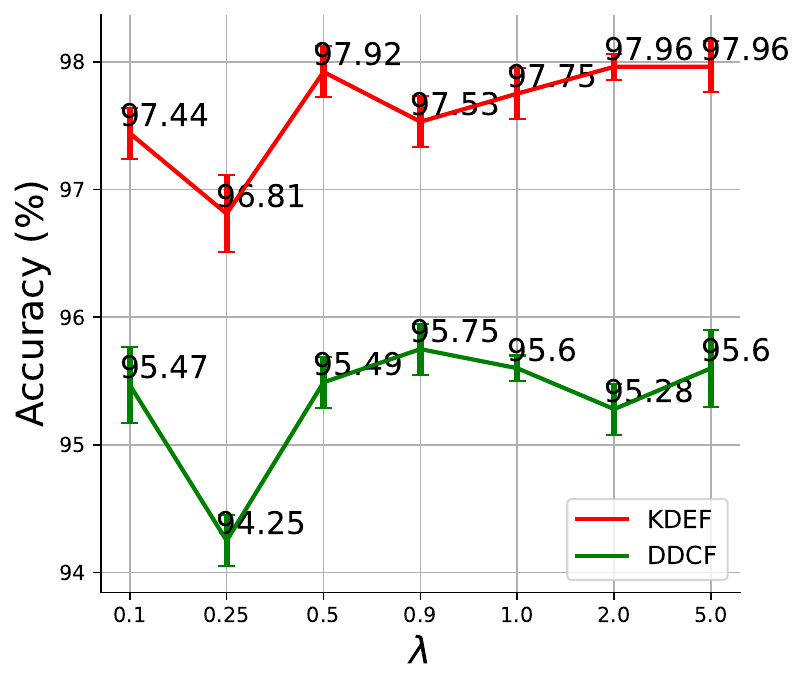}
         \caption{Acc. vs. $\lambda$ for \textit{\textbf{ReMixMatch}}.}
        \label{fig:small_lambda_remixmatch}
     \end{subfigure}\\
     \begin{subfigure}[b]{0.15\textwidth}
         \centering
         \includegraphics[width=1.1\textwidth]{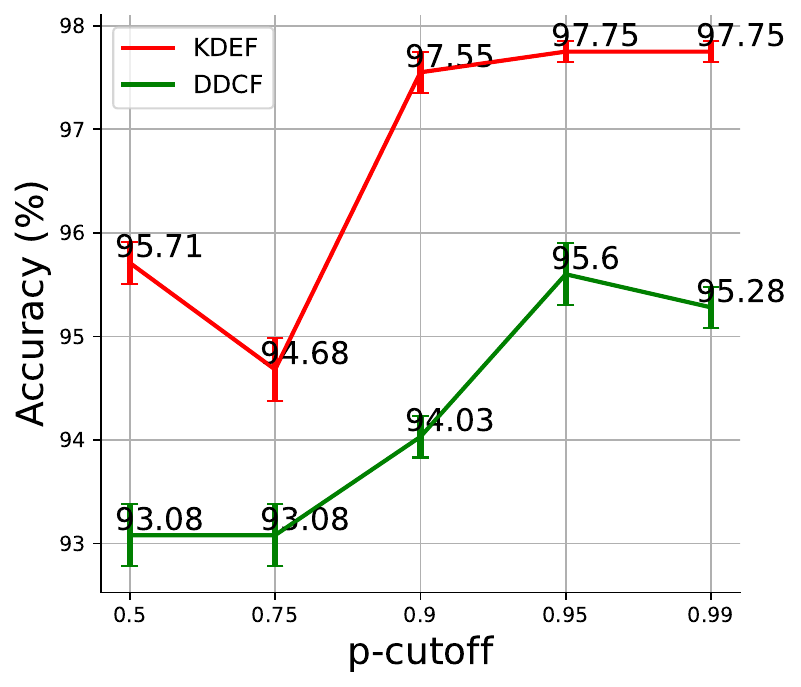}
         \caption{Acc. vs. $P_\text{cutoff}$ for \textit{\textbf{FixMatch}}.}
         \label{fig:small_cutoff_ccssl}
     \end{subfigure}
     \hspace{5pt}
      \begin{subfigure}[b]{0.15\textwidth}
         \centering
         \includegraphics[width=1.1\textwidth]{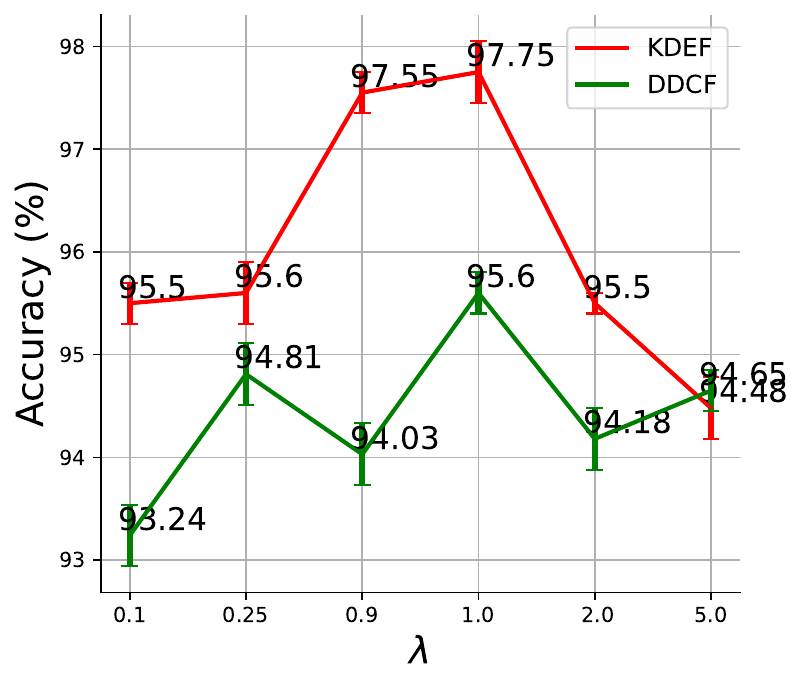}
         \caption{Acc. vs. $\lambda$ for \textit{\textbf{FixMatch}}.}
         \label{fig:small_lambda_ccssl}
     \end{subfigure}

    \caption{Sensitivity study of various parameters for two of the best semi-supervised methods on small unlabelled data. }
    \label{fig:sensitivity_small}
\end{figure}

\begin{figure*}[ht]
    \centering

     \begin{subfigure}[b]{0.18\textwidth}
         \centering
         \includegraphics[width=1.1\textwidth]{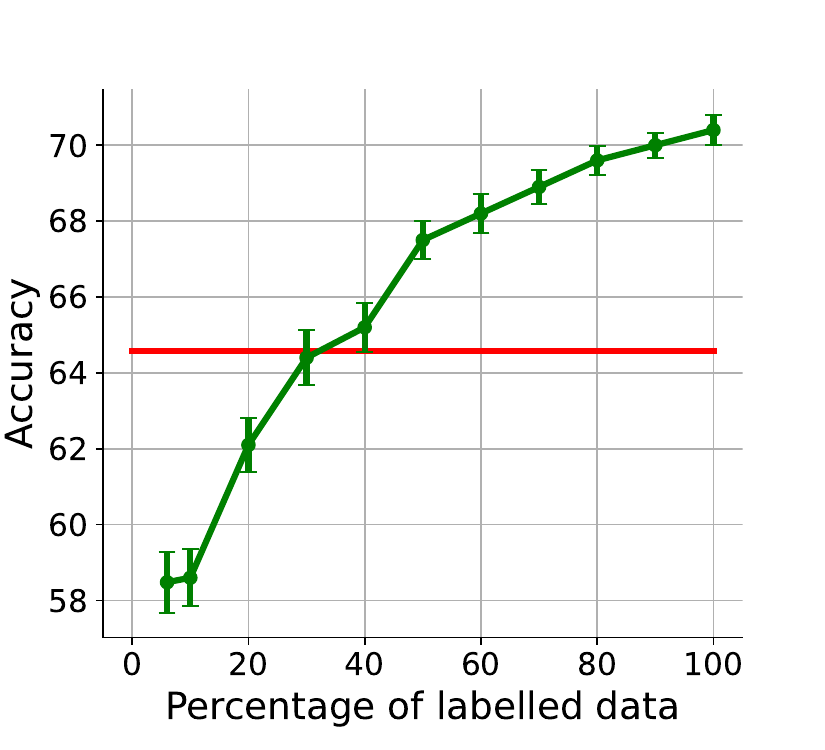}
         \caption{FER13 with OOD.}
         \label{fig:fer_ood_cross_sup}
     \end{subfigure}
     \begin{subfigure}[b]{0.18\textwidth}
         \centering
         \includegraphics[width=1.1\textwidth]{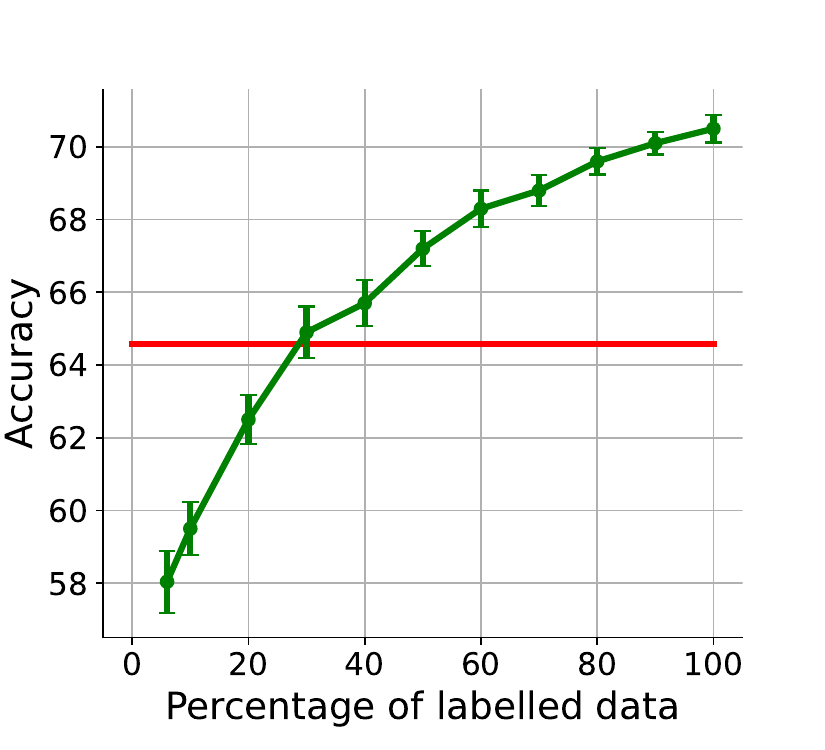}
         \caption{FER13 with Unc.}
        \label{fig:fer_unc_cross_sup}
     \end{subfigure}
     \begin{subfigure}[b]{0.18\textwidth}
         \centering
         \includegraphics[width=1.1\textwidth]{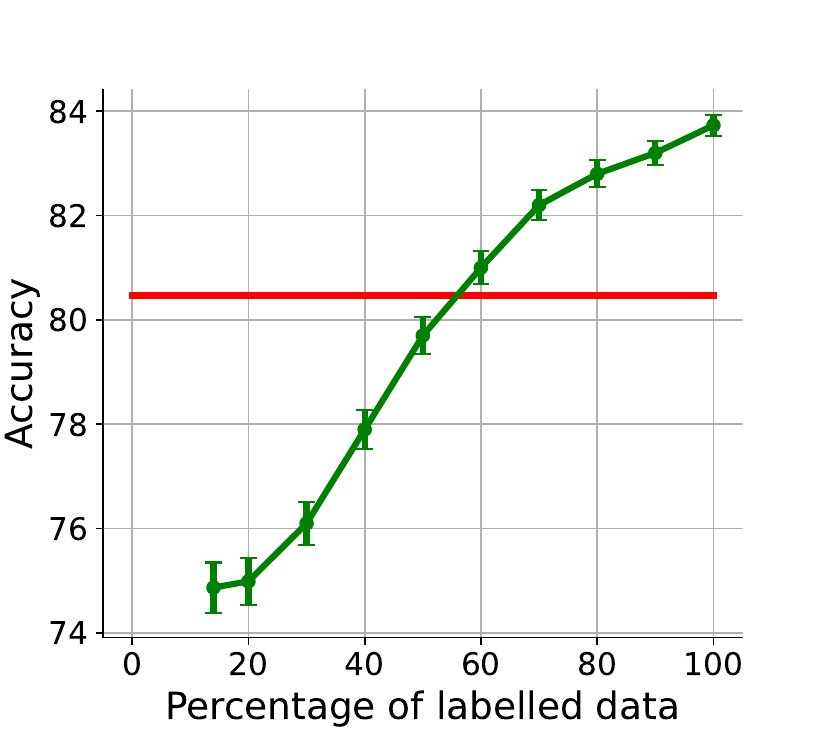}
         \caption{RAF-DB with OOD.}
         \label{fig:raf_ood_cross_sup}
     \end{subfigure}
      \begin{subfigure}[b]{0.18\textwidth}
         \centering
         \includegraphics[width=1.1\textwidth]{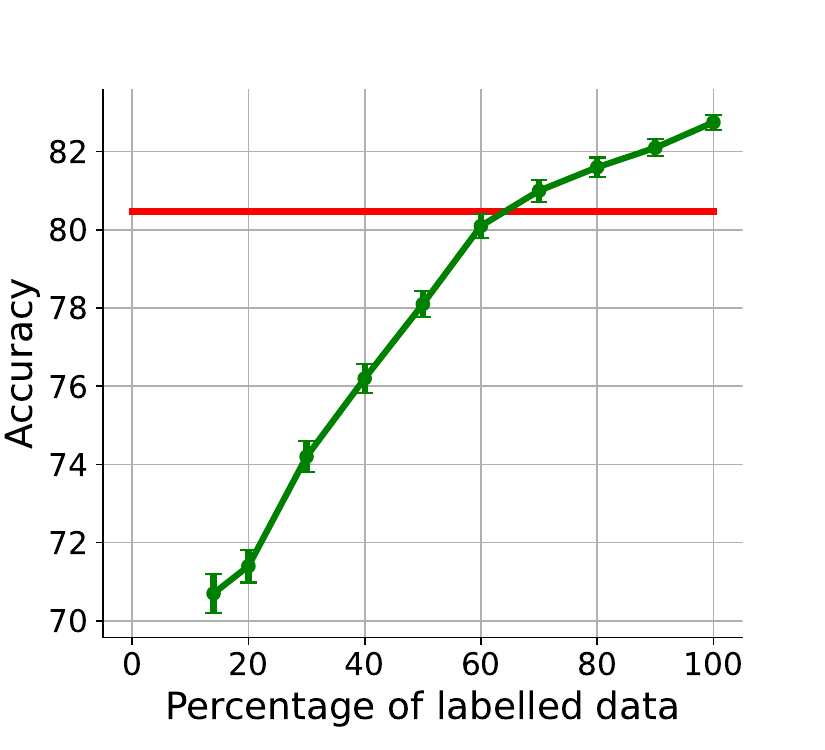}
         \caption{RAF-DB with Unc.}
         \label{fig:raf_unc_cross_sup}
     \end{subfigure}
     \begin{subfigure}[b]{0.18\textwidth}
         \centering
         \includegraphics[width=1.1\textwidth]{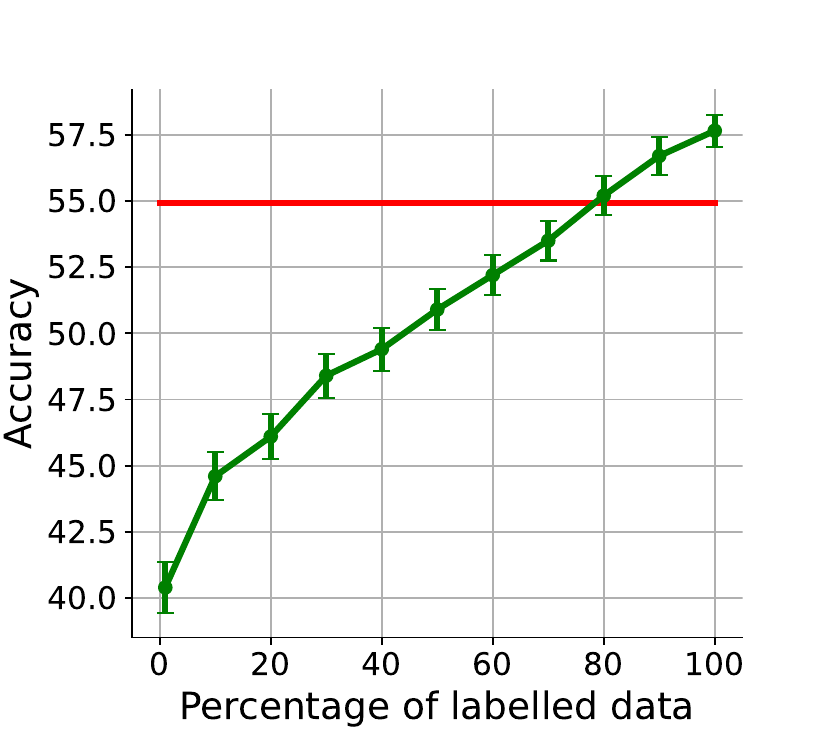}
         \caption{AffectNet with Unc.}
         \label{fig:affectnet_unc_cross_sup}
     \end{subfigure}

    \caption{Performance for different percentages of labelled data. Red line indicates fully-supervised learning with all samples. }
    \label{fig:labelled_data_perc}
\end{figure*}

\subsubsection{Sensitivity Study}
Next, we present a sensitivity study on the main hyper-parameters of the two best methods for unconstrained semi-supervised learning (ReMixMatch and MixMatch). The results are presented in Figure \ref{fig:sensitivity_unconstrained}.

The first experiment on ReMixMatch's $\alpha$ values shows that different datasets have different optimal values. Specifically, the optimal values for RAF-DB, FER13, and AffectNet are 0.5, 0.75, and 0.1, respectively (Figure \ref{fig:unc_alpha_remixmatch}). Similarly, optimal values for $\lambda$ experiments are different for each dataset (Figure \ref{fig:unc_lambda_remixmatch}). In this case, the optimal values for RAF-DB, FER13, and AffectNet are 2.0, 1.0, and 2.0, respectively. MixMatch experiments follow a similar pattern of having different optimal values. For $\alpha$, the best values are 0.5, 0.75, and 0.1 (Figure \ref{fig:unc_alpha_mixmatch}), and for $\lambda$ (Figure \ref{fig:unc_lambda_mixmatch}), the optimal values are 0.5, 2.0, and 0.5 for RAF-DB, FER13, and AffectNet, respectively. In conclusion, hyper-parameters are generally specific to each dataset when learning from unconstrained unlabelled data.

\subsubsection{Discussion}
Table \ref{tab:unconstrained_summary} summarizes the results of learning from unconstrained unlabelled data. The key takeaway from this table is that although semi-supervised learning with unconstrained data achieves lower performance compared to ID semi-supervised learning, it still outperforms fully supervised learning when an equal number of labelled samples are used. For instance, FER13 achieves an accuracy of 53.58\% with fully supervised learning, while the same dataset reaches 58.04\% accuracy through the utilization of semi-supervised learning with unconstrained unlabelled data. \textcolor{black}{We also report the results for using the whole dataset as the labelled data, and unconstrained unlabelled data. The results from this study show that using semi-supervised learning with unconstrained unlabelled data considerably improves performance over fully-supervised learning. Here, the improvements are 5.93\%, 2.28\%, and 2.73\% for FER-13, RAF-DB, and AffectNet, respectively.}

\subsection{Semi-supervised FER with small ID unlabelled data}

\subsubsection{Setup}
\textcolor{black}{In this section, we discuss the results of semi-supervised learning with small ID unlabelled data. While gathering substantial quantities of unlabelled ID data can pose challenges, obtaining a smaller set of unlabelled ID data might be more feasible in some cases. Consequently, we also conduct experiments with small but ID unlabelled data for completeness. To this end, we utilize small datasets collected in lab environments. More specifically, we conduct the experiments on KDEF \cite{kdef} and DDCF \cite{ddcf} datasets, and present the results for a similar number of labelled samples as in previous experiments, with the remaining samples being used as the unlabelled set. Performances are reported on the validation set of the corresponding datasets. }

\subsubsection{Performance}
Table \ref{tab:small_dataset} shows the main results of different semi-supervised methods in this setting. ReMixMatch again performs the best across all methods. With ID data in this setting, FixMatch again performs well and shows the second-best average accuracy. The average accuracy for ReMixMatch and FixMatch are 73.09\% and 71.76\%, respectively.

\subsubsection{Sensitivity study} 
We also analyze the sensitivity of the two best-performing methods, ReMixMatch and FixMatch, in this setting. Figure \ref{fig:sensitivity_small} illustrates the results of this study. The experiment on the $\alpha$ parameter of ReMixMatch (Figure \ref{fig:small_alpha_remixmatch}) shows that higher values lead to better accuracy. The optimal values are found to be 0.9 and 0.75 for DDCF and KDEF datasets, respectively. Similarly, for the $\lambda$ parameter (Figure \ref{fig:small_lambda_remixmatch}), the best results are obtained for 0.9 and 2.0 for DDCF and KDEF datasets. On the other hand, the experiments on FixMatch reveal that the default parameters of the original FixMatch yield the best accuracy for both datasets. Specifically, a $P_\text{cutoff}$ value of 0.95 (Figure \ref{fig:small_cutoff_ccssl}) and a $\lambda$ value of 1.0 (Figure \ref{fig:small_lambda_ccssl}) achieve the best performances.

\subsection{General discussions}
\textcolor{black}{
In this section, we provide a discussion on some of our key observations. More specifically, we present a comparison between different unlabelled settings of SSL, a comparison to supervised learning, and a few insights for improving SSL for FER.}

\textcolor{black}{
\subsubsection{Comparing different unlabelled settings}
To better understand the overall behaviour of SSL using different categories of unlabeled data (ID vs. OOD vs. unconstrained), we present the performance of the \textit{best-performing} method for each setting on FER-13 in Table \ref{tab:discussion}. The results demonstrate that expectedly, the highest performance (62.20\%) can be achieved with ID unlabelled data when a limited number of labelled samples (250) are available. We also observe that with limited OOD and unconstrained unlabelled data, similar performances can be achieved, underperforming the ID setting.
Interestingly, the results show that when the amount of unlabelled data is considerably scaled, unconstrained unlabelled data are in fact more beneficial for SSL, in comparison to ID unlabelled data. Moreover, the similarity between OOD and unconstrained performance persists. These results indicate that using free-living data irrespective of their potential data or class distributions is a viable and effective approach for SSL when sufficiently large amounts of unlabelled data can be collected.
}

\textcolor{black}{
\subsubsection{Comparison to supervised learning}
To understand the impact of the amount of labelled data on SSL, we perform a detailed analysis by using different percentages of labelled data. As seen in Figure \ref{fig:labelled_data_perc}, we expectedly observe that more labelled data help the model in learning better representations. However, the figure shows that the added benefit of using more labelled samples decreases as more and more labelled samples are incorporated. Additionally, in Figure \ref{fig:labelled_data_perc}, we present the performance of fully-supervised learning (shown in the figures with a red line) in comparison to SSL. An interesting observation from this analysis is that irrespective of the type of unlabelled data used, SSL always outperforms fully supervised learning given a sufficient number of labelled samples. This further demonstrates the effectiveness of SSL with unconstrained/OOD data and the viability of reducing reliance on labels given the availability of unlabelled data.}

\textcolor{black}{
\subsubsection{Insights for improving SSL}
While applying existing SSL methods to FER, we noticed that directly transferring their default setups for this purpose does not yield optimal performance. To improve the performance of existing methods, we conduct a comprehensive study to identify FER-specific best practices for various aspects of the SSL methods. Furthermore, we find the different methods perform differently in different semi-supervised settings studied in the paper. Overall, our study reveals the following key insights into semi-supervised learning for FER. First, when learning from more challenging learning scenarios, such as OOD, unconstrained, and small unlabelled data, the unsupervised loss plays a more critical role. Thus, increasing the value of the unlabelled loss factor ($\lambda$) improves performance. 
Secondly, considering the important role of unlabeled data and acknowledging that large volumes of such data can greatly enhance performance, we foresee a new perspective toward developing SSL for FER. Specifically, we anticipate that integrating \textit{strong} unsupervised methods with small/modest supervised frameworks can result in robust and generalized frameworks, leading to the creation of more scalable SSL techniques in the area.}

\begin{table}[]
\caption{\textcolor{black}{Comparison across different settings on FER-13. }}
\begin{center}
\begin{tabular}{>{\color{black}}l | >{\color{black}}c >{\color{black}}c}
\toprule
 & \multicolumn{2}{c}{\textbf{Accuracy (\%) }} \\
\textbf{Setup} &  \textbf{250 labels} & \textbf{All}  \\ \midrule
No unlabelled data & 53.58\textcolor{black}{\tiny $\pm 1.1$} & 64.57\textcolor{black}{\tiny $\pm 0.9$}\\ 
SSL (ID) &  \textbf{62.20}\textcolor{black}{\tiny $\pm 0.5$} & 65.15\textcolor{black}{\tiny $\pm 0.5$} \\
SSL (OOD) & 58.48\textcolor{black}{\tiny $\pm 0.8$} & {70.40}\textcolor{black}{\tiny $\pm 0.8$} \\ 
SSL (Unconstrained) & 58.04\textcolor{black}{\tiny $\pm 0.8$} & \textbf{70.50}\textcolor{black}{\tiny $\pm 0.4$} \\

\bottomrule
\end{tabular}
\label{tab:discussion}
\end{center}
\end{table}

\section{Conclusion} 
This research offers a comprehensive analysis of 11 semi-supervised methods for FER. The study evaluates the performance of these methods in various unlabelled data scenarios, including ID, OOD, unconstrained, and very small unlabelled sets. Our primary finding is that FixMatch is the most effective semi-supervised method for learning from ID unlabelled data. However, for all other real-world scenarios (OOD, unconstrained, and small set), ReMixMatch consistently outperforms other semi-supervised methods. Another noteworthy finding is that semi-supervised learning from any data scenario produces better results in comparison to fully-supervised learning from the same number of labelled samples. When learning from an ID unlabelled set, semi-supervised methods can produce a performance improvement of up to 11\% over the fully-supervised method. Although compared to ID, performance is generally lower for both OOD and unconstrained unlabelled data, these methods still outperform fully-supervised learning. Interestingly, the unconstrained setting, despite being significantly more challenging than ID or OOD, underperforms OOD by only a small margin. This is a significant observation since collecting large amounts of unconstrained unlabelled data is considerably easier and more practical than collecting ID or even OOD (but constrained) data. Overall, we anticipate that this research will serve as a useful guide for further investigation into semi-supervised learning in the context of FER, as well as other domains.
\textcolor{black}{One limitation of this study is that we primarily focused on facial expression datasets with static images of macro-expression. Exploring the effectiveness of semi-supervised learning for dynamic or micro expressions remains an interesting area for future research. While our study highlights the promise of unconstrained unlabelled data on the overall performance of semi-supervised FER, further investigation is needed to understand the impact of data quality and potential biases within such datasets.}

\section*{Acknowledgements}
We would like to thank BMO Bank of Montreal and Mitacs for funding this research. We are also thankful to SciNet HPC Consortium for helping with the computation resources.

\bibliographystyle{IEEEtran}
\bibliography{ref.bib}

\begin{thebibliography}{10}
\providecommand{\url}[1]{#1}
\csname url@samestyle\endcsname
\providecommand{\newblock}{\relax}
\providecommand{\bibinfo}[2]{#2}
\providecommand{\BIBentrySTDinterwordspacing}{\spaceskip=0pt\relax}
\providecommand{\BIBentryALTinterwordstretchfactor}{4}
\providecommand{\BIBentryALTinterwordspacing}{\spaceskip=\fontdimen2\font plus
\BIBentryALTinterwordstretchfactor\fontdimen3\font minus \fontdimen4\font\relax}
\providecommand{\BIBforeignlanguage}[2]{{%
\expandafter\ifx\csname l@#1\endcsname\relax
\typeout{** WARNING: IEEEtran.bst: No hyphenation pattern has been}%
\typeout{** loaded for the language `#1'. Using the pattern for}%
\typeout{** the default language instead.}%
\else
\language=\csname l@#1\endcsname
\fi
#2}}
\providecommand{\BIBdecl}{\relax}
\BIBdecl

\bibitem{kolahdouzi2022facetoponet}
M.~Kolahdouzi, A.~Sepas-Moghaddam, and A.~Etemad, ``Facetoponet: Facial expression recognition using face topology learning,'' \emph{IEEE Transactions on Artificial Intelligence}, 2022.

\bibitem{kolahdouzi2021face}
------, ``Face trees for expression recognition,'' in \emph{IEEE International Conference on Automatic Face and Gesture Recognition}, 2021, pp. 1--5.

\bibitem{tokuno2011usage}
S.~Tokuno, G.~Tsumatori, S.~Shono, E.~Takei, T.~Yamamoto, G.~Suzuki, S.~Mituyoshi, and M.~Shimura, ``Usage of emotion recognition in military health care,'' in \emph{Defense Science Research Conference and Expo}, 2011, pp. 1--5.

\bibitem{thrasher2011mood}
M.~Thrasher, M.~D. Van~der Zwaag, N.~Bianchi-Berthouze, and J.~H. Westerink, ``Mood recognition based on upper body posture and movement features,'' in \emph{International Conference on Affective Computing and Intelligent Interaction}, 2011, pp. 377--386.

\bibitem{sanchez2013inferring}
D.~Sanchez-Cortes, J.-I. Biel, S.~Kumano, J.~Yamato, K.~Otsuka, and D.~Gatica-Perez, ``Inferring mood in ubiquitous conversational video,'' in \emph{12th International Conference on Mobile and Ubiquitous Multimedia}, 2013, pp. 1--9.

\bibitem{leng2007experimental}
H.~Leng, Y.~Lin, and L.~Zanzi, ``An experimental study on physiological parameters toward driver emotion recognition,'' in \emph{International Conference on Ergonomics and Health Aspects of Work with Computers}, 2007, pp. 237--246.

\bibitem{pseudo_labels}
D.-H. Lee \emph{et~al.}, ``Pseudo-label: The simple and efficient semi-supervised learning method for deep neural networks,'' in \emph{Workshop on Challenges in Representation Learning, ICML}, vol.~3, no.~2, 2013, p. 896.

\bibitem{vat}
T.~Miyato, S.-i. Maeda, M.~Koyama, and S.~Ishii, ``Virtual adversarial training: a regularization method for supervised and semi-supervised learning,'' \emph{IEEE Transactions on Pattern Analysis and Machine Intelligence}, vol.~41, no.~8, pp. 1979--1993, 2018.

\bibitem{uda}
Q.~Xie, Z.~Dai, E.~Hovy, T.~Luong, and Q.~Le, ``Unsupervised data augmentation for consistency training,'' \emph{Advances in Neural Information Processing Systems}, vol.~33, pp. 6256--6268, 2020.

\bibitem{fixmatch}
K.~Sohn, D.~Berthelot, N.~Carlini, Z.~Zhang, H.~Zhang, C.~A. Raffel, E.~D. Cubuk, A.~Kurakin, and C.-L. Li, ``Fixmatch: Simplifying semi-supervised learning with consistency and confidence,'' \emph{Advances in Neural Information Processing Systems}, vol.~33, pp. 596--608, 2020.

\bibitem{remixmatch}
D.~Berthelot, N.~Carlini, E.~D. Cubuk, A.~Kurakin, K.~Sohn, H.~Zhang, and C.~Raffel, ``Remixmatch: Semi-supervised learning with distribution matching and augmentation anchoring,'' in \emph{International Conference on Learning Representations}, 2020.

\bibitem{yoshihashi2019classification}
R.~Yoshihashi, W.~Shao, R.~Kawakami, S.~You, M.~Iida, and T.~Naemura, ``Classification-reconstruction learning for open-set recognition,'' in \emph{IEEE/CVF Conference on Computer Vision and Pattern Recognition}, 2019, pp. 4016--4025.

\bibitem{guo2020safe}
L.-Z. Guo, Z.-Y. Zhang, Y.~Jiang, Y.-F. Li, and Z.-H. Zhou, ``Safe deep semi-supervised learning for unseen-class unlabeled data,'' in \emph{International Conference on Machine Learning}, 2020, pp. 3897--3906.

\bibitem{UnMixMatch}
S.~Roy and A.~Etemad, ``Scaling up semi-supervised learning with unconstrained unlabelled data,'' in \emph{AAAI Conference on Artificial Intelligence}, 2024.

\bibitem{auxmix}
A.~Banitalebi-Dehkordi, P.~Gujjar, and Y.~Zhang, ``Auxmix: semi-supervised learning with unconstrained unlabeled data,'' in \emph{IEEE/CVF Conference on Computer Vision and Pattern Recognition}, 2022, pp. 3999--4006.

\bibitem{acii2022}
S.~Roy and A.~Etemad, ``Analysis of semi-supervised methods for facial expression recognition,'' in \emph{IEEE International Conference on Affective Computing and Intelligent Interaction}, 2022, pp. 1--8.

\bibitem{pi_model}
M.~Sajjadi, M.~Javanmardi, and T.~Tasdizen, ``Regularization with stochastic transformations and perturbations for deep semi-supervised learning,'' \emph{Advances in Neural Information Processing Systems}, vol.~29, 2016.

\bibitem{mean_teacher}
A.~Tarvainen and H.~Valpola, ``Mean teachers are better role models: Weight-averaged consistency targets improve semi-supervised deep learning results,'' \emph{Advances in Neural Information Processing Systems}, vol.~30, 2017.

\bibitem{mixmatch}
D.~Berthelot, N.~Carlini, I.~Goodfellow, N.~Papernot, A.~Oliver, and C.~A. Raffel, ``Mixmatch: A holistic approach to semi-supervised learning,'' \emph{Advances in Neural Information Processing Systems}, vol.~32, 2019.

\bibitem{flexmatch}
B.~Zhang, Y.~Wang, W.~Hou, H.~Wu, J.~Wang, M.~Okumura, and T.~Shinozaki, ``Flexmatch: Boosting semi-supervised learning with curriculum pseudo labeling,'' \emph{Advances in Neural Information Processing Systems}, vol.~34, pp. 18\,408--18\,419, 2021.

\bibitem{comatch}
J.~Li, C.~Xiong, and S.~C. Hoi, ``Comatch: Semi-supervised learning with contrastive graph regularization,'' in \emph{IEEE/CVF International Conference on Computer Vision}, 2021, pp. 9475--9484.

\bibitem{ccssl}
F.~Yang, K.~Wu, S.~Zhang, G.~Jiang, Y.~Liu, F.~Zheng, W.~Zhang, C.~Wang, and L.~Zeng, ``Class-aware contrastive semi-supervised learning,'' in \emph{IEEE/CVF Conference on Computer Vision and Pattern Recognition}, 2022, pp. 14\,421--14\,430.

\bibitem{fer13}
I.~J. Goodfellow, D.~Erhan, P.~L. Carrier, A.~Courville, M.~Mirza, B.~Hamner, W.~Cukierski, Y.~Tang, D.~Thaler, D.-H. Lee \emph{et~al.}, ``Challenges in representation learning: A report on three machine learning contests,'' in \emph{International Conference on Neural Information Processing}, 2013, pp. 117--124.

\bibitem{raf_db}
S.~Li, W.~Deng, and J.~Du, ``Reliable crowdsourcing and deep locality-preserving learning for expression recognition in the wild,'' in \emph{IEEE/CVF Conference on Computer Vision and Pattern Recognition}, 2017, pp. 2852--2861.

\bibitem{affectnet}
A.~Mollahosseini, B.~Hasani, and M.~H. Mahoor, ``Affectnet: A database for facial expression, valence, and arousal computing in the wild,'' \emph{IEEE Transactions on Affective Computing}, vol.~10, no.~1, pp. 18--31, 2017.

\bibitem{kdef}
D.~Lundqvist, A.~Flykt, and A.~{\"O}hman, ``The karolinska directed emotional faces (kdef),'' \emph{CD ROM from Department of Clinical Neuroscience, Psychology Section, Karolinska Institutet}, vol.~91, no. 630, pp. 2--2, 1998.

\bibitem{ddcf}
K.~A. Dalrymple, J.~Gomez, and B.~Duchaine, ``The dartmouth database of children’s faces: Acquisition and validation of a new face stimulus set,'' \emph{PloS One}, vol.~8, no.~11, p. e79131, 2013.

\bibitem{celeba}
Z.~Liu, P.~Luo, X.~Wang, and X.~Tang, ``Deep learning face attributes in the wild,'' in \emph{Proceedings of International Conference on Computer Vision}, December 2015.

\bibitem{noisystudent}
Q.~Xie, M.-T. Luong, E.~Hovy, and Q.~V. Le, ``Self-training with noisy student improves imagenet classification,'' in \emph{IEEE/CVF Conference on Computer Vision and Pattern Recognition}, 2020, pp. 10\,687--10\,698.

\bibitem{MPL}
H.~Pham, Z.~Dai, Q.~Xie, and Q.~V. Le, ``Meta pseudo labels,'' in \emph{IEEE/CVF Conference on Computer Vision and Pattern Recognition}, 2021, pp. 11\,557--11\,568.

\bibitem{randaugment}
E.~D. Cubuk, B.~Zoph, J.~Shlens, and Q.~V. Le, ``Randaugment: Practical automated data augmentation with a reduced search space,'' in \emph{IEEE/CVF Conference on Computer Vision and Pattern Recognition Workshops}, 2020, pp. 702--703.

\bibitem{augmix}
D.~Hendrycks, N.~Mu, E.~D. Cubuk, B.~Zoph, J.~Gilmer, and B.~Lakshminarayanan, ``Augmix: A simple data processing method to improve robustness and uncertainty,'' in \emph{International Conference on Learning Representations}, 2020.

\bibitem{mixup}
H.~Zhang, M.~Cisse, Y.~N. Dauphin, and D.~Lopez-Paz, ``mixup: Beyond empirical risk minimization,'' in \emph{International Conference on Learning Representations}, 2018.

\bibitem{kurup2019semi}
A.~R. Kurup, M.~Ajith, and M.~M. Ram{\'o}n, ``Semi-supervised facial expression recognition using reduced spatial features and deep belief networks,'' \emph{Neurocomputing}, vol. 367, pp. 188--197, 2019.

\bibitem{cohen2003semi}
I.~Cohen, N.~Sebe, F.~G. Cozman, and T.~S. Huang, ``Semi-supervised learning for facial expression recognition,'' in \emph{5th ACM SIGMM international workshop on Multimedia information retrieval}, 2003, pp. 17--22.

\bibitem{li2022towards}
H.~Li, N.~Wang, X.~Yang, X.~Wang, and X.~Gao, ``Towards semi-supervised deep facial expression recognition with an adaptive confidence margin,'' in \emph{IEEE/CVF Conference on Computer Vision and Pattern Recognition}, 2022, pp. 4166--4175.

\bibitem{PT}
J.~Jiang and W.~Deng, ``Boosting facial expression recognition by a semi-supervised progressive teacher,'' \emph{IEEE Transactions on Affective Computing}, 2021.

\bibitem{Rethink-Self-SSL}
B.~Fang, X.~Li, G.~Han, and J.~He, ``Rethinking pseudo-labeling for semi-supervised facial expression recognition with contrastive self-supervised learning,'' \emph{IEEE Access}, 2023.

\bibitem{CFRN}
H.~Sun, C.~Pi, and W.~Xie, ``Semi-supervised facial expression recognition by exploring false pseudo-labels,'' in \emph{IEEE International Conference on Multimedia and Expo}, 2023, pp. 234--239.

\bibitem{autoaugment}
E.~D. Cubuk, B.~Zoph, D.~Mane, V.~Vasudevan, and Q.~V. Le, ``Autoaugment: Learning augmentation strategies from data,'' in \emph{IEEE/CVF Conference on Computer Vision and Pattern Recognition}, 2019, pp. 113--123.

\bibitem{rafdb}
S.~Li, W.~Deng, and J.~Du, ``Reliable crowdsourcing and deep locality-preserving learning for expression recognition in the wild,'' in \emph{IEEE/CVF Conference on Computer Vision and Pattern Recognition}, 2017, pp. 2852--2861.

\bibitem{resnet}
K.~He, X.~Zhang, S.~Ren, and J.~Sun, ``Deep residual learning for image recognition,'' in \emph{IEEE/CVF Conference on Computer Vision and Pattern Recognition}, 2016, pp. 770--778.

\bibitem{oliver2018realistic}
A.~Oliver, A.~Odena, C.~A. Raffel, E.~D. Cubuk, and I.~Goodfellow, ``Realistic evaluation of deep semi-supervised learning algorithms,'' \emph{Advances in Neural Information Processing Systems}, 2018.

\bibitem{su2021realistic}
J.-C. Su, Z.~Cheng, and S.~Maji, ``A realistic evaluation of semi-supervised learning for fine-grained classification,'' in \emph{IEEE/CVF Conference on Computer Vision and Pattern Recognition}, 2021, pp. 12\,966--12\,975.

\end{thebibliography}

\begin{IEEEbiography}[{\includegraphics[width=1in,height=1.25in,clip,keepaspectratio]{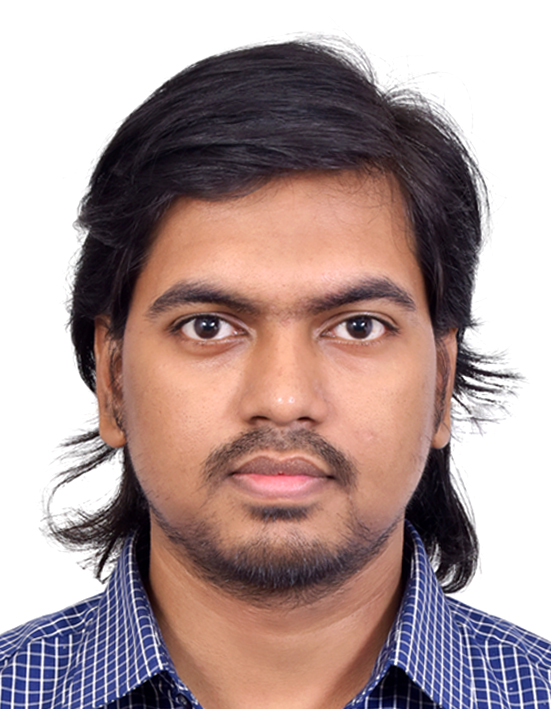}}]{Shuvendu Roy}
is a Ph.D. student at Department of Electrical and Computer Engineering, and Ingenuity Labs Research Institute, at Queen’s University in Canada. He received his B.Sc. in Computer Science and Engineering from Khulna University of Engineering \& Technology, Bangladesh. His current research is focused on computer vision with deep learning, as well as self-supervised and semi-supervised learning. 
\end{IEEEbiography}
\vspace{-25pt}
\begin{IEEEbiography}[{\includegraphics[width=1in,height=1.25in,clip,keepaspectratio]{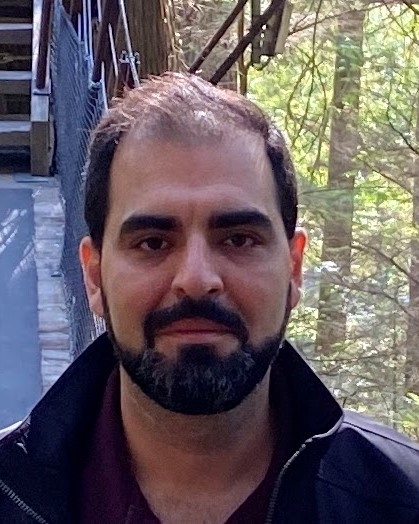}}]{Ali Etemad}
is an Associate Professor at the Department of Electrical and Computer Engineering, Queen's University. He holds an endowed professership of Mitchell Professor in AI for Human Sensing \& Understanding. He leads the Ambient Intelligence and Interactive Machines (Aiim) lab. He received his M.A.Sc. and Ph.D. degrees in Electrical and Computer Engineering from Carleton University, Ottawa, Canada, in 2009 and 2014, respectively.  His main areas of research are machine learning and deep learning focused on human-centered applications with wearables, smart devices, and smart environments. Prior to joining Queen’s, he held several industrial positions as lead scientist. He has published over 160 papers in top venues in the area, is a co-inventor of 10 patents, and has given over 25 invited talks at different venues.  Dr. Etemad is an Associate Editor for IEEE Transactions on Affective Computing and IEEE Transactions on Artificial Intelligence. He has served as a PC member/reviewer, and has held organizing roles at various venues. He has received a number of awards including Supervisor of the Year Award (at Queen’s), Instructor of the Year Award (at Queen’s), and several Best Paper Awards (e.g., at ACM ICMI'23). Dr. Etemad’s lab and research program have been funded by the Natural Sciences and Engineering Research Council (NSERC) of Canada, Ontario Centers of Excellence (OCE), Canadian Foundation for Innovation (CFI), Mitacs, and other organizations, as well as the private sector.
\end{IEEEbiography}

\end{document}